\documentclass{article}

\usepackage{colm2024_conference}

\usepackage{graphicx}
\usepackage[utf8]{inputenc} %
\usepackage[T1]{fontenc}    %
\usepackage{hyperref}       %
\usepackage{url}            %
\usepackage{booktabs}       %
\usepackage{amsfonts}       %
\usepackage{nicefrac}       %
\usepackage{microtype}      %
\usepackage{xcolor}     %
\usepackage{colortbl}
\usepackage{wrapfig}
\usepackage{marvosym}
\usepackage{algorithm}
\usepackage{algorithmic}
\usepackage{booktabs}
\usepackage{multirow}
\usepackage[table]{xcolor}
\usepackage{subcaption}
\usepackage{titletoc}

\makeatletter
\newcommand\blfootnote[1]{%
  \begingroup
  \renewcommand\thefootnote{}\footnote{#1}%
  \addtocounter{footnote}{-1}%
  \endgroup
}
\makeatother

\definecolor{darkred}{RGB}{177, 38, 26}
\definecolor{darkblue}{RGB}{67, 116, 177}
\definecolor{darkgreen}{rgb}{0.0, 0.5, 0.0}

\definecolor{bestcol}{RGB}{  0,102,204} %
\definecolor{goodcol}{RGB}{ 34,139, 34} %
\definecolor{deltaBg}{RGB}{220,230,255} %

\usepackage[inline]{enumitem}

\usepackage[most]{tcolorbox}
\usepackage{etoc}

\tcbset{
  colback=white, %
  colframe=black, %
  arc=2mm,            %
  boxrule=1pt,      %
  left=6pt, right=6pt, top=6pt, bottom=6pt  %
}

\usepackage{epigraph}
\usepackage{amsmath, amssymb}
\usepackage[capitalize,noabbrev]{cleveref}

\usepackage{multirow}
\usepackage{array} 

\newlist{inlinelist}{enumerate*}{1}
\setlist[inlinelist]{label=(\roman*)}

\title{Deeper is Not Always Better: Mitigating the Alignment Tax via Confident Layer Decoding}

\author{%
 \textbf{Xuanming Zhang*$^{1}$, Sining Zhoubian*$^{2}$, Yuxuan Chen$^{1}$, Tianyi Tang$^{1}$, An Yang$^{1}$, Sean Du$^{3}$, Chujie Zheng$^{1}$, Fei Huang$^{1}$, Dayiheng Liu$^{1}$, Gao Huang$^{2}$, Jingren Zhou$^{1}$} \\
 $^1$ Qwen Team, Alibaba Inc. $^2$ Tsinghua University $^3$ Nanyang Technological University 
}

\begin{document}

\blfootnote{Emails: \texttt{xuemuqiangu@gmail.com}\quad\texttt{zbsn21@mails.tsinghua.edu.cn}}
\blfootnote{$^*$Equal contribution.}

\maketitle

\begin{abstract}

Autoregressive generation in large language models (LLMs) conventionally decodes from the final layer, assuming that deeper representations yield more reliable next-token predictions. We revisit this assumption by revealing a recurring \emph{Guess--Refine--Perturb} dynamic: early layers form coarse guesses, intermediate layers refine reasoning-relevant semantics, and final layers can perturb these refined predictions toward generic or alignment-preferred tokens. We introduce \emph{Confident Decoding}, a training-free decoding strategy that dynamically selects the most reliable near-final layer through entropy-guided conservative backward search. We further provide a theoretical formulation of layer selection as an optimal stopping problem, showing that under bounded projection noise and dominant late-stage alignment perturbation, our search rule filters perturbation while bounding the loss relative to the oracle refinement layer. Experiments across dense and Mixture-of-Experts LLMs demonstrate consistent gains on challenging reasoning benchmarks, including GPQA-Diamond, Omni-MATH, and HLE, with zero memory overhead and less than 2\% latency increase. These results suggest dynamically bypassing final-layer perturbations can unlock stronger reasoning behavior from aligned LLMs.

\begin{figure}[h]
    \centering
    \includegraphics[width=0.68\linewidth]{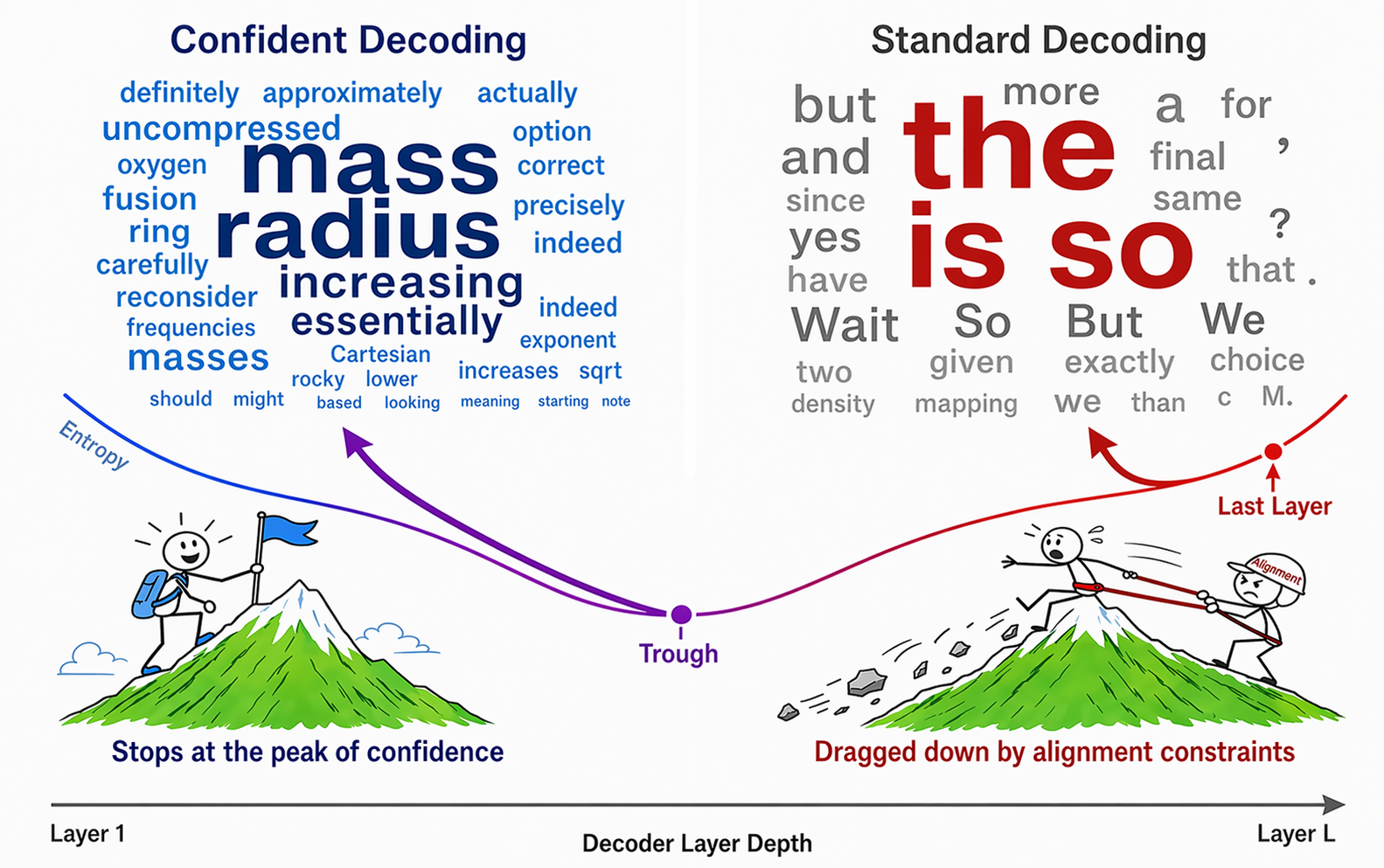}
    \caption{Token substitutions produced by Confident Decoding on Qwen3.5-35B-A3B. When the two decoding strategies disagree ($\sim$2\% of generated tokens), Standard Decoding (\textbf{right}) selects generic, high-frequency function words and punctuation (e.g., ``the'', ``is'', ``so'', ``.'') characteristic of alignment-induced common-word bias in the final layers. Confident Decoding (\textbf{left}) instead commits at the entropy valley---the layer of peak confidence before late-stage perturbation---and recovers domain-specific, semantically precise terminology (e.g., ``mass'', ``radius'', ``approximately'', ``Cartesian''). Word sizes are proportional to substitution frequency. The central curve illustrates the predictive entropy $H(p^{(l)}_t)$ along decoder layer depth $l$. The ``Trough'' marks the entropy valley where the model's internal confidence peaks; beyond this point, alignment constraints in deeper layers drag predictions toward safe but uninformative continuations.}
    \label{fig:token_substitution}
\end{figure}
\end{abstract}

\section{Introduction} \label{sec: introduction}

Autoregressive generation in large language models (LLMs) conventionally decodes each token from the final decoder layer. This standard practice implicitly assumes that representations become progressively more reliable with depth: shallow layers provide incomplete computations, intermediate layers refine contextual information, and the deepest layer produces the most accurate next-token distribution~\citep{vaswani2017attention,shazeer2017outrageously,gupta2025llms}. Under this view, the final layer is treated as the natural and optimal interface between the model's internal computation and the output vocabulary.

However, recent evidence suggests that this monotonic-depth assumption can break down. Probing and layer-wise analyses show that intermediate layers often encode strong task-relevant semantics, while later layers may compress, redirect, or perturb information that has already been refined~\citep{skean2025layer,csordaslanguage}. Through a detailed analysis of residual-stream dynamics, contribution norms, and prediction trajectories, we identify a recurring \emph{Guess--Refine--Perturb} pattern in LLM forward passes (see \S\ref{subsec: phase-definition} and Figure~\ref{fig: layer wise}). Early layers produce coarse statistical guesses; intermediate layers refine these guesses along a stable semantic trajectory; and, for a subset of tokens, the final layers introduce a sharp representational shift that can move the prediction away from the most reasoning-relevant intermediate state.

We hypothesize that this late-stage perturbation is partly related to the objectives used in modern post-training. Contemporary LLMs are typically adapted through continued pre-training, supervised fine-tuning, and preference-alignment procedures such as RLHF, RLAIF, or DPO~\citep{zixuan2023continual,ouyang2022training,lee2024rlaif,rafailov2023direct}. These procedures improve instruction following, safety, and response style, but they may also encourage final-layer distributions to favor frequent, safe, or generic continuations~\citep{askell2021general,zou2023representation}. For ordinary conversational or safety-oriented prompts, such alignment behavior can serve as a useful guardrail. For complex reasoning tasks, however, the same late-stage bias may conflict with the task-specific reasoning trajectory formed in intermediate layers, producing what we call a \emph{planning--pragmatics tradeoff}: the model may internally form a strong reasoning-oriented prediction, but the final layer may pragmatically shift the output distribution toward a more generic or alignment-preferred token.

Several recent decoding strategies have begun to exploit intermediate-layer information. Contrastive decoding methods, such as Decoding by Contrasting Layers~\citep{chuangdola,zhou2025alw}, compare predictions across layers to suppress undesirable continuations, while logits-evolution methods study how token distributions change through depth~\citep{zhang2024sled,das2025entropy,zhang2025cognition}. Early-exit methods also use intermediate confidence signals to reduce inference cost~\citep{schuster2022confident,yang2025dynamic}. Yet these approaches either continue to treat the final layer as the main semantic anchor or focus primarily on efficiency rather than reasoning degradation. They do not directly ask whether the final layer should always be the layer from which the next-token distribution is emitted.

To address this question, we propose \emph{Confident Decoding}, a training-free, drop-in decoding strategy that dynamically selects the most reliable near-final layer at each generation step. Importantly, Confident Decoding does not truncate the transformer or modify the model's forward pass. The only difference is that, instead of always forwarding final-layer logits to the sampler, Confident Decoding computes candidate logits from a small near-final layer window and selects the first local entropy trough encountered by a conservative backward search. Since lower token entropy corresponds to a sharper predictive distribution, this rule identifies the point at which the model reaches high confidence before a potential late-layer perturbation emerges (see Figure~\ref{fig:token_substitution} for a representative example).

We evaluate Confident Decoding across a broad range of challenging benchmarks, including general reasoning, mathematical problem solving, long-context understanding, coding, and safety, with datasets such as GPQA-Diamond~\citep{rein2024gpqa}, HLE~\citep{center2026benchmark}, Omni-MATH~\citep{gaoomni}, LongBench v2~\citep{bai2025longbench}, LiveCodeBench v6~\citep{jainlivecodebench}, and Air-Bench-2024~\citep{zeng2025air}. Across dense and Mixture-of-Experts architectures, Confident Decoding consistently improves over standard greedy decoding and strong contrastive baselines, while introducing negligible memory overhead and less than 2\% additional latency. We further observe that larger and stronger models can exhibit more pronounced late-layer perturbations, making dynamic layer selection increasingly beneficial as model capability and task difficulty grow.

In summary, our contributions are threefold:
\begin{itemize}[leftmargin=10pt, topsep=0pt,itemsep=3pt]
    \item We identify a recurring \emph{Guess--Refine--Perturb} dynamic in LLM forward passes, showing that final-layer representations are not always the most reliable source for reasoning-sensitive next-token prediction.
    \item We propose \emph{Confident Decoding}, a training-free and drop-in decoding strategy that preserves the full forward pass while dynamically selecting near-final logits through entropy-guided conservative backward search.
    \item We provide theoretical (\S\ref{sec: theoretical-grounding}) and empirical (\S\ref{sec: experiment}) evidence that Confident Decoding filters late-layer perturbations, preserves refined reasoning signals, and improves performance across diverse reasoning, coding, long-context, and safety benchmarks.
\end{itemize}

\section{Preliminaries} \label{sec: preliminaries}

In this section, we formally characterize the forward-pass dynamics of LLMs to motivate our proposed decoding strategy. We first quantify the layer-wise representational shifts to define the three-phase progression mathematically. Subsequently, we present a pilot study validating the superiority of dynamic, entropy-based layer selection over static exit paradigms.

\subsection{Layer-wise Dynamics} \label{subsec: phase-definition}

\begin{figure}[!t]
    \centering
    \begin{minipage}[t]{0.47\linewidth}
        \centering
        \includegraphics[width=\linewidth]{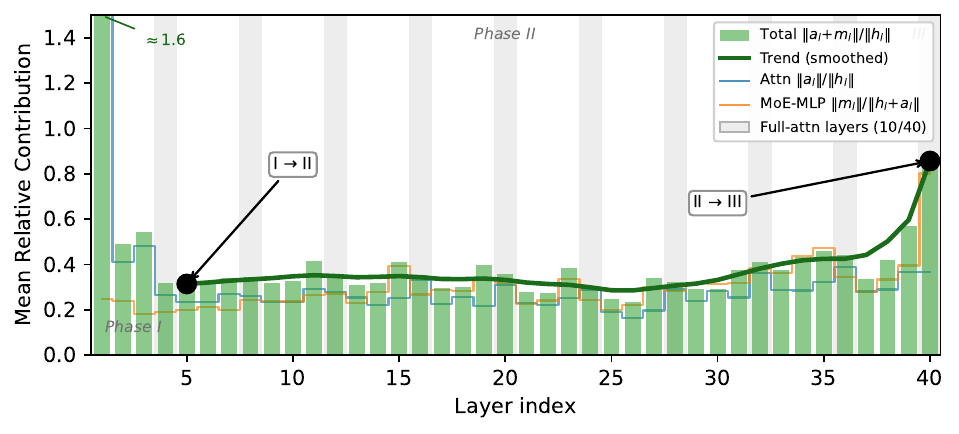}
        \subcaption{Relative Contribution Norm}
        \label{fig: layer wise norm}
    \end{minipage}
    \hfill
    \begin{minipage}[t]{0.47\linewidth}
        \centering
        \includegraphics[width=\linewidth]{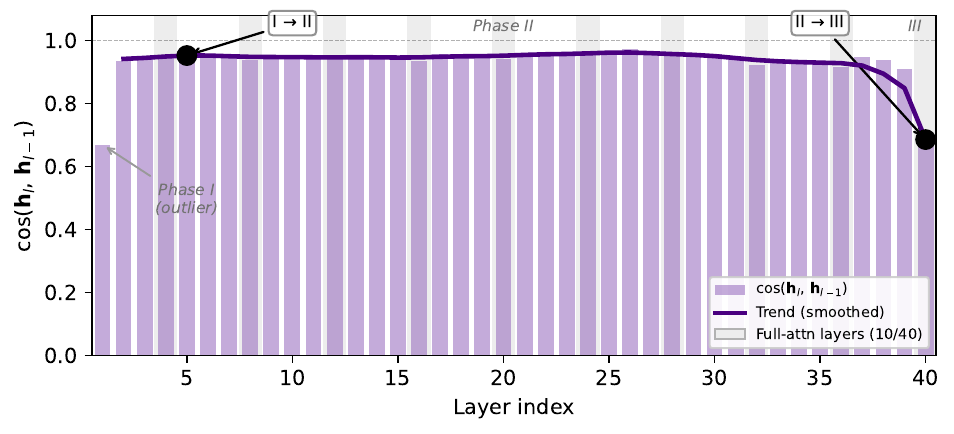}
        \subcaption{Residual I/O Cosine Similarity}
        \label{fig: layer wise cossim}
    \end{minipage}
    \caption{Layer-wise dynamics of Qwen3.5-35B-A3B on GSM8K~\citep{cobbe2021training}. Gray bands mark the 10 full-attention layers
    (at $l = 4, 8, \ldots, 40$) which enable global token interactions.
    \textbf{(a) Relative Contribution Norm}: The first decoder
    layer's update dominates the embedding magnitude; contributions stabilize through
    Phase~II; the final full-attention layer ($l\!=\!40$) resurges markedly in Phase~III.
    \textbf{(b) Residual I/O Cosine Similarity}: IO-CosSim remains high throughout Phase~II,
    confirming directionally faithful refinement, before dropping sharply at $l\!=\!40$---the largest
    directional deflection in the Phase~II--III regime.}
    \label{fig: layer wise}
\end{figure}

Consider an $L$-layer auto-regressive Transformer-based language model. Let $\mathcal{V}$ denote the vocabulary space. For a given input sequence up to step $t$, the residual stream at layer $l$, denoted as $\mathbf{h}^{(l)}_t \in \mathbb{R}^d$, is updated via the attention and feed-forward sub-layers. In the standard sequential (pre-norm) formulation:
\begin{align}
\tilde{\mathbf{h}}^{(l)}_t &= \mathbf{h}^{(l-1)}_t + f_{\text{Attn}}^{(l)}(\mathbf{h}^{(l-1)}_t), \label{eq:attn-update}\\
\mathbf{h}^{(l)}_t &= \tilde{\mathbf{h}}^{(l)}_t + f_{\text{FFN}}^{(l)}(\tilde{\mathbf{h}}^{(l)}_t), \label{eq:ffn-update}
\end{align}
where $f_{\text{Attn}}^{(l)}$ denotes the attention sublayer (full or linear) and $f_{\text{FFN}}^{(l)}$ the feed-forward sublayer (MLP or MoE). We define the \textit{layer contribution vector} as the total update applied by layer~$l$:
$$\mathbf{m}^{(l)}_t \;\triangleq\; \mathbf{h}^{(l)}_t - \mathbf{h}^{(l-1)}_t \;=\; f_{\text{Attn}}^{(l)}(\mathbf{h}^{(l-1)}_t) + f_{\text{FFN}}^{(l)}(\tilde{\mathbf{h}}^{(l)}_t)$$
so that $\mathbf{h}^{(l)}_t = \mathbf{h}^{(l-1)}_t + \mathbf{m}^{(l)}_t$. This definition naturally extends to parallel-attention variants where both sublayers receive $\mathbf{h}^{(l-1)}_t$ directly. To understand how each layer alters the semantic trajectory of the token prediction, we analyze two key metrics across network depth: the \textit{Relative Contribution Norm} and the \textit{Residual I/O Cosine Similarity}~\citep{skean2025layer,csordaslanguage}.

$$\text{Norm Ratio}^{(l)} = \frac{\|\mathbf{m}^{(l)}_t\|_2}{\|\mathbf{h}^{(l-1)}_t\|_2}, \quad \text{IO-CosSim}^{(l)} = \frac{\mathbf{h}^{(l)}_t \cdot \mathbf{h}^{(l-1)}_t}{\|\mathbf{h}^{(l)}_t\|_2 \|\mathbf{h}^{(l-1)}_t\|_2}$$

Under the residual stream decomposition $\mathbf{h}^{(l)}_t = \mathbf{h}^{(l-1)}_t + \mathbf{m}^{(l)}_t$, $\text{Norm Ratio}^{(l)}$ characterizes the \textit{write intensity}. When $\text{Norm Ratio}^{(l)} \gg 1$, the contribution dominates the existing state ($\mathbf{h}^{(l)}_t \approx \mathbf{m}^{(l)}_t$), effectively overwriting the accumulated representation; when $\text{Norm Ratio}^{(l)} \ll 1$, the layer applies incremental corrections without displacing prior content; and when $\text{Norm Ratio}^{(l)} \lesssim 1$, the new write is comparable in magnitude to the carried residual~\citep{elhage2021mathematical}. In this intermediate regime, $\text{Norm Ratio}^{(l)}$ alone cannot distinguish constructive reinforcement from disruptive rewriting, because the net effect depends on the update direction. $\text{IO-CosSim}^{(l)}$ therefore characterizes \textit{directional fidelity}: in the high-dimensional representation space where direction encodes semantic content, a value near 1 indicates the layer preserves and refines the existing semantic trajectory, while lower values indicate rotation into a semantically distinct subspace~\citep{mikolov2013distributed}.

As shown in Figure~\ref{fig: layer wise}, empirical observations on Qwen3.5-35B-A3B---a model interleaving 30 linear-attention (DeltaNet) layers with 10 full-attention layers at fixed positions (shaded in gray)---reveal a consistent three-phase structure, which we term the \textit{Guess-Refine-Perturbation} progression:

\begin{enumerate}[leftmargin=15pt, topsep=2pt, itemsep=2pt]
\item \textbf{Phase I: Guess (Shallow Layers, $l \lesssim 0.15L$).} In the initial layers, $\text{Norm Ratio}^{(l)}$ is exceedingly high: $\text{Norm Ratio}^{(1)} \approx 1.6$, meaning the first decoder layer's contribution vector is $1.6$ times the magnitude of the incoming embedding, so the output $\mathbf{h}^{(1)}_t$ is almost entirely determined by the layer's computation rather than the token embedding itself. Correspondingly, $\text{IO-CosSim}^{(1)} \approx 0.67$---markedly below the Phase~II plateau---indicating that the model undergoes a substantial directional shift in this phase, rapidly constructing an initial latent representation under high uncertainty.


\item \textbf{Phase II: Refine (Intermediate Layers, $0.15L \lesssim l \lesssim 0.95L$).} $\text{Norm Ratio}^{(l)}$ drops sharply and stabilizes in the range $0.23$--$0.57$ ($< 1$): the sublayer contribution is substantially smaller than the existing residual, so each layer performs incremental, directionally faithful updates that progressively integrate contextual information without displacing the accumulated representation. Correspondingly, $\text{IO-CosSim}^{(l)}$ remains consistently high ($0.91$--$0.97$) throughout, confirming that the semantic trajectory is refined but not rotated. The model refines its token predictions along a stable semantic trajectory.

\item \textbf{Phase III: Perturbation (Post Layers, $l \gtrsim 0.95L$).} In the final layers, $\text{Norm Ratio}^{(l)}$ re-elevates markedly above the Phase~II plateau, peaking at $\text{Norm Ratio}^{(40)}$--- $2$--$3\times$ the Phase~II level. This value remains below~$1$ and thus does \emph{not} enter the Phase~I overwriting regime. They indicate a write whose magnitude is comparable to the incoming residual rather than a small corrective update. The corroborating evidence comes from $\text{IO-CosSim}^{(40)} \approx 0.69$---equivalently, an output-state rotation relative to the incoming residual---which falls well below the Phase~II band ($0.91$--$0.97$) and marks the sharpest directional deflection outside the initial embedding phase. The combination of these two metrics indicates a structurally significant, directionally misaligned update that partially rewrites and biases the semantic trajectory refined during Phase~II. We formalize the mechanism behind this perturbation in \S\ref{subsec: alignment-tax}.\end{enumerate}

This Phase III perturbation breaks the monotonic improvement assumption, suggesting that extracting predictions at the end of Phase II may yield superior reasoning accuracy over decoding from the final layer~\citep{gupta2025llms}. This three-phase pattern is further validated on additional architectures in Appendix~\ref{app:9b-layerwise}.

\subsection{Motivation: Emerging Entropy Valley in Intermediate Layer} \label{subsec: trough-experiment}

Prior studies and observations in \S\ref{subsec: phase-definition} suggest that LLMs often converge on stable semantic representations before Phase III~\citep{skean2025layer}. This motivates us to seek a reliable intermediate output point or proxy to bypass potential late-stage perturbations without the overhead of retraining auxiliary classifiers. 

\begin{figure}[!t]
    \centering
    \includegraphics[width=.95\linewidth]{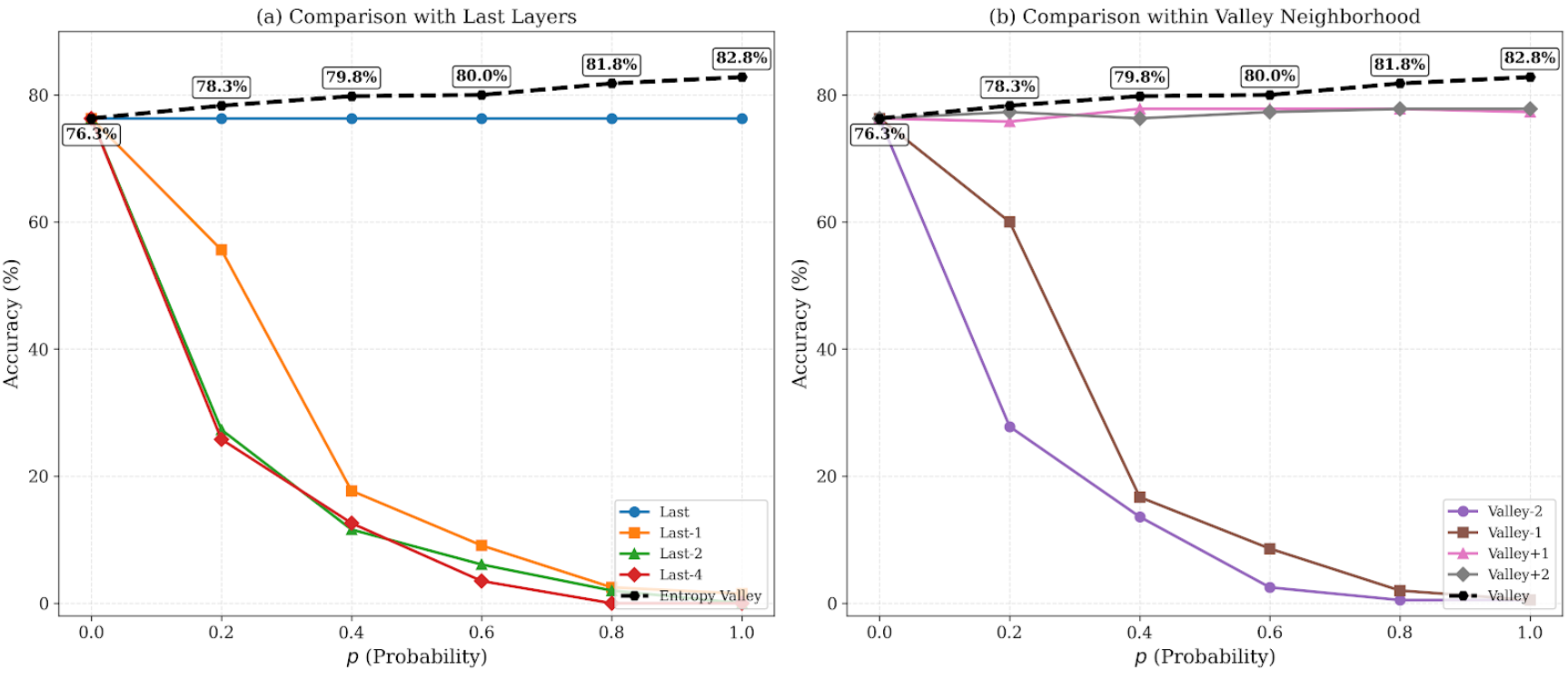}
    \caption{Comparison of layer-selection strategies on GPQA-Diamond, using Qwen3.5-35B-A3B as the base model. Dynamic valley selection consistently outperforms last-layer decoding and static early exits, while selecting dynamic layers neighboring the valley also degrades performance. This supports our central claim that the optimal decoding depth is token-dependent and typically lies near the Phase~II/III boundary rather than at a fixed layer, and the entropy valley provides a confidence boundary for optimized performance.}
    \label{fig: superiority}
\end{figure}

\textbf{Limitations of Static Early Exit.} We first investigate the performance of a naive static early exit strategy. As shown in Figure~\ref{fig: superiority}(a), an independent Bernoulli probability $p$ represents the likelihood of overriding the final layer and forcing decoding from a fixed shallower layer $(L-k)$ at any given token generation step. As this execution probability $p$ increases, the overall model accuracy undergoes a precipitous decline. This observation indicates that static truncation ignores the inherent variance in token complexity; a uniform exit policy prematurely interrupts the essential computation required for ``hard'' tokens, thereby destroying the model's reasoning integrity. Consequently, a robust extraction mechanism must be \textit{token-adaptive}, requiring a dynamic indicator to identify the optimal depth for each hidden state.

\textbf{Entropy as a Confidence Metric.} A natural candidate for such an indicator is the model's predictive confidence. By applying the pre-trained language modeling head $W_U$ to intermediate hidden states $\mathbf{h}_t^{(l)}$, we can elicit a layer-wise probability distribution $p_t^{(l)}$. To ensure dimensional consistency and avoid softmax collapse, we apply the final-layer normalization before projection:
\begin{equation}
    p_t^{(l)} = \text{Softmax}\left(W_U \cdot \text{RMSNorm}_L(\mathbf{h}_t^{(l)})\right)
\end{equation}
The uncertainty at each layer is then quantified by the Shannon entropy $H(p_t^{(l)}) = -\sum p_t^{(l)} \log p_t^{(l)}$.

We notice that using the original $W_U$ as a zero-shot probe introduces a \textit{basis shift} problem~\citep{belrose2023eliciting}, as $W_U$ is explicitly optimized for the final layer's latent space. While representation homogeneity in the latter half of modern LLMs mitigates this~\citep{wang2025understanding}, the mapping error still increases as we move further away from the final layer. This creates a trade-off: deeper layers offer smaller mapping bias, while certain intermediate layers might host more ``confident'' or ``pure'' semantic information before alignment-induced noise or over-smoothing occurs.

To balance mapping reliability with prediction confidence, we propose the \textbf{Entropy Valley}---the local entropy minimum encountered when scanning backwards from the final layer. Intuitively, this valley represents a state where the model has reached a consensus on the output distribution before the final layers potentially introduce detrimental oscillations. 

\textbf{Empirical Validation.} Figure~\ref{fig: superiority} illustrates the superiority of this dynamic selection. Unlike static exits, the \textit{Entropy Valley} strategy maintains high accuracy even when applied frequently. Furthermore, Figure~\ref{fig: superiority}(b) demonstrates that the valley itself is a precise optimal boundary: decoding from its immediate neighbors (Valley$\pm k$) leads to consistent performance degradation. These results confirm that the entropy valley serves as a reliable, token-dependent proxy for the zenith of the refinement phase, justifying its use as the core boundary for our proposed rollback mechanism.

\section{Theoretical Grounding} \label{sec: theoretical-grounding}

To formalize the superiority of Confident Decoding, we analyze the forward-pass dynamics through the lens of Information Theory and Representation Engineering. We demonstrate why the representational optimum emerges at an intermediate depth and prove that our conservative backward search acts as a mathematically optimal, adaptive filter that selectively neutralizes alignment perturbations while preserving safety guardrails.

\subsection{Information-Theoretic View of Layer Dynamics} \label{subsec: info-bottleneck}

While the Information Bottleneck (IB) principle~\citep{tishby2015deep} typically characterizes neural network training, the learned weight matrices internalize this optimization, dictating the layer-wise information flow during inference. Let $X$ represent the input context and $Y_{\text{logic}}$ represent the optimal target token strictly derived from domain-specific reasoning. The residual stream $\mathbf{h}^{(l)}_t$ evolves by navigating the IB tradeoff:
$$\mathcal{L}_{\text{IB}} = I(X; \mathbf{h}^{(l)}_t) - \beta I(\mathbf{h}^{(l)}_t; Y_{\text{logic}})$$

\begin{figure}[!t]
    \centering
    \begin{minipage}[t]{0.48\linewidth}
        \centering
        \includegraphics[width=\linewidth]{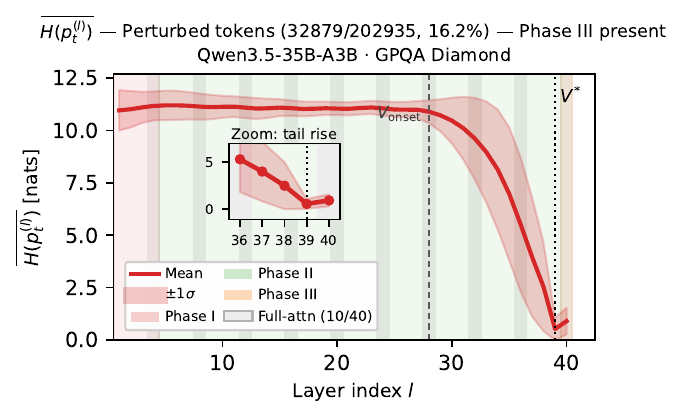}
        \subcaption{Perturbed tokens (16.2\%) --- Phase~III present}
        \label{fig: theory rising}
    \end{minipage}
    \hfill
    \begin{minipage}[t]{0.48\linewidth}
        \centering
        \includegraphics[width=\linewidth]{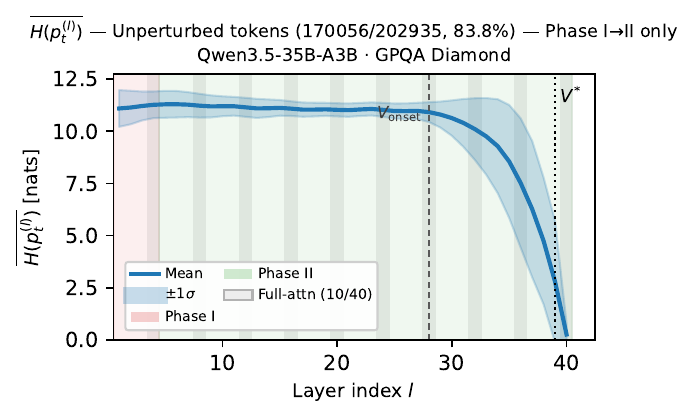}
        \subcaption{Unperturbed tokens (83.8\%) --- no Phase~III}
        \label{fig: theory falling}
    \end{minipage}
    \caption{Mean logit-lens entropy $\overline{H(p^{(l)}_t)}$ per layer for Qwen3.5-35B-A3B on GPQA Diamond ($N\!=\!50$ prompts, 4{,}096 generated tokens per prompt, 202{,}935 tokens total).
    \textbf{(a) Perturbed tokens (16.2\%, $\Delta H\!=\!{+0.37}$ nats):} Tokens already at low entropy at $V^*$ ($\bar{H}_{V^*}\!=\!0.52$ nats). The final full-attention layer introduces an upward perturbation, the alignment-tax signature, disrupting a nearly committed prediction. These tokens undergo Phase~III.
    \textbf{(b) Unperturbed tokens (83.8\%, $\Delta H\!=\!{-2.52}$ nats):} Content tokens still uncertain at $V^*$ ($\bar{H}_{V^*}\!=\!2.78$ nats). The final layer continues its Phase~II refinement, driving entropy toward zero with no perturbation. The backward scan exploits this heterogeneity: it selects $V^*$ for perturbed tokens (bypassing Phase~III) and $L$ for unperturbed ones (utilizing full refinement). The same per-token heterogeneity is confirmed on additional models in Appendix~\ref{app:9b-entropy}.}
    \label{fig: theory prove}
\end{figure}

During \textbf{Phase I (Guess)}, the network aggressively compresses the superficial input $X$ (minimizing $I(X; \mathbf{h})$). This severe dimensionality reduction induces geometric volatility, rendering the predictive entropy $H(p^{(l)}_t)$ highly unstable and susceptible to degenerate unigram biases. Critically, the residual influence of this compression noise carries over into the early portion of Phase~II, leaving the predictive entropy on a high-entropy plateau that is empirically indistinguishable from Phase~I, as shown in Figure~\ref{fig: theory prove}.

The network then transitions into \textbf{Phase II (Refine)}, where successive attention mechanisms rigorously integrate contextual dependencies, dominating the flow to maximize $I(\mathbf{h}^{(l)}_t; Y_{\text{logic}})$. Within Phase~II, we identify an internal inflection layer $V_{\text{onset}}$ at which the carry-over of Phase I dissipates and the network enters its monotone refinement regime. Ideally, within this restricted domain $l \ge V_{\text{onset}}$, the predictive entropy $H(p^{(l)}_t)$ serves as an empirical upper bound for the true conditional entropy $H(Y_{\text{logic}} | \mathbf{h}^{(l)}_t)$. As mutual information monotonically increases across this late-refine window, $H(p^{(l)}_t)$ follows a strict monotone downward trajectory. 

The minimum at $V^{*}$ represents the exact representational zenith before potential perturbation sets in. As shown in Figure~\ref{fig: theory prove}(a), on Qwen3.5-35B-A3B over GPQA Diamond generating 4{,}096 tokens per prompt, Phase~II accumulates $\Delta H \approx -10.8$ nats across 202{,}935 tokens ($V_{\rm onset}\!\approx\!28$, $V^*\!=\!39$). Crucially, Phase~III is not a fixed architectural region but a \emph{per-token} phenomenon determined by whether the final layer perturbs a token's prediction: 16.2\% of tokens exhibit an entropy rise of $+0.37$ nats at $l\!=\!40$--- these tokens undergo Phase~III. Figure~\ref{fig:token_substitution} shows examples of such tokens being replaced by their lower-entropy alternatives from an earlier layer. The remaining 83.8\% of tokens still resolving at $V^*$ show a further entropy drop of $-2.52$ nats---these tokens simply complete their normal Phase~I$\to$II refinement through the final layer.

\subsection{Modeling Alignment: Tax vs. Guardrail} \label{subsec: alignment-tax}

Modern LLMs undergo rigorous post-training to align with human preferences. Recent findings in Representation Engineering~\citep{zou2023representation} reveal that such alignment behaviors are governed by low-rank steering vectors, predominantly activated in the final layers to modulate outputs without overwriting pre-trained knowledge. Thus, in \textbf{Phase III (Perturbation)}, the latent representations optimize a regularized risk toward a generic, safe distribution $P_{\text{align}}$:
$$\mathcal{R}^{(l)} = \mathbb{E} [-\log p^{(l)}_t(Y_{\text{logic}}|X)] + \lambda \mathcal{D}_{\text{KL}}(P_{\text{logic}}^{(l)} \| P_{\text{align}})$$
where $\lambda$ governs the steering intensity. Let $V^*$ be the oracle transition boundary between Phase II and Phase III. For $l \to L$, the network induces an additive perturbation $\delta^{(l)}_{\text{align}}$ to minimize the KL divergence: $\mathbf{h}^{(l)}_t \approx \mathbf{h}^{(V^*)}_t + \sum_{k=V^*+1}^{l} \delta^{(k)}_{\text{align}}$.

Crucially, the destructive nature of this perturbation is strictly conditional:
\begin{itemize}[leftmargin=15pt, topsep=2pt, itemsep=2pt]
    \item \textbf{Alignment as a Guardrail (Safety Tasks):} For standard conversational or safety queries, $P_{\text{logic}} \approx P_{\text{align}}$. The KL divergence is negligible. The late-stage layers refine formatting without semantic disruption.
    \item \textbf{Alignment as a Tax (Complex Reasoning):} For domains like rigorous mathematics and science, the specialized logic distribution $P_{\text{logic}}$ sharply conflicts with the generic $P_{\text{align}}$. The additive perturbation forcibly steers the latent state away from the reasoning subspace, mathematically manifesting as an unnatural "entropy oscillation" that derails the established logic chain.
\end{itemize}

\subsection{Minimax Optimality of Conservative Backward Search} \label{subsec: backward-optimality}

Empirical evidence from prior work consistently shows that, beyond roughly the network midpoint, late-layer hidden states become approximately linearly decodable through the frozen unembedding $W_U$~\citep{belrose2023eliciting,elhoushi2024layerskip,park2024linear}; since our empirical $V_{\mathrm{onset}}$ falls well within this regime ($V_{\mathrm{onset}} \gtrsim 0.5L$), the projection noise $\epsilon^{(l)}$ for $l \ge V_{\mathrm{onset}}$ is correspondingly small and bounded by a tight $\epsilon_{\max}$.

We formulate dynamic layer selection as an Optimal Stopping Problem. The observable entropy $\hat{H}(l)$ decomposes into:
$$\hat{H}(l) = H^*(l) + \epsilon^{(l)} + \eta^{(l)}$$
where $H^*(l)$ is the monotonic true entropy ($\Delta H^*(l) \le 0$ for $l \in [V_{\text{onset}}, V^*]$), $\epsilon^{(l)}$ is bounded projection noise ($|\epsilon^{(l)}| \le \epsilon_{\max}$), and $\eta^{(l)}$ is the per-token alignment perturbation ($\eta^{(l)} \to 0$ for $l \le V^*$; for tokens where the alignment correction conflicts with the task-optimal prediction, $\Delta \eta^{(l)} > 2\epsilon_{\max}$ in the final layers).

\textbf{Theorem 1 (Minimax Optimality).} \textit{Let $\hat{V} = \max \{ l < L \mid \hat{H}(l-1) \ge \hat{H}(l) \}$. The conservative backward scan strictly guarantees $\hat{V} \in [V_{\text{onset}}, V^*]$, effectively filtering alignment perturbations while bounding semantic precision loss.}

\textbf{Proof.} 

\textit{1. Adaptive Evasion of Phase III Tax:} For tokens where $\Delta \eta^{(l)} > 2\epsilon_{\max} > |\Delta \epsilon^{(l)}|$, the observed gradient for $l > V^*$ is strictly positive ($\Delta \hat{H}(l) > 0$). Thus, the stopping condition is never triggered in Phase III, guaranteeing $\hat{V} \le V^*$ and nullifying the alignment tax ($\eta^{(\hat{V})} \to 0$). Conversely, for tokens where the alignment correction is synergistic with refinement, $\Delta \eta^{(l)} \to 0$, preventing false triggers and naturally halting at $\hat{V} \approx L$.

\textit{2. Bounded Optimality in Phase II:} The scan enters Phase II at $l = V^*$. Here, $\eta^{(l)} \to 0$. The stopping condition $\hat{H}(\hat{V}-1) \ge \hat{H}(\hat{V})$ is triggered. We evaluate two exhaustive scenarios:

\begin{itemize}[leftmargin=15pt, topsep=2pt, itemsep=2pt]\item \textbf{Case A (Strong Integration Signal):} If the semantic refinement is strong such that $|\Delta H^*(V^*)| > |\Delta \epsilon^{(V^*)}|$, then the true signal overcomes the projection noise, yielding $\hat{H}(V^*-1) > \hat{H}(V^*)$. The algorithm stops exactly at $\hat{V} = V^*$. The selection is oracle-optimal.\item \textbf{Case B (Weak Integration / Local Oscillation):} Suppose at some layer $k \le V^*$, the integration signal weakens ($\Delta H^*(k) \to 0$), and projection noise induces a micro-oscillation where $\hat{H}(k-1) < \hat{H}(k)$. The algorithm prematurely stops at $\hat{V} = k$. However, because this only occurs when $\Delta H^*$ is exceedingly small, the semantic information $I(\mathbf{h}^{(k)}; Y) \approx I(\mathbf{h}^{(V^*)}; Y)$. The semantic loss $\mathcal{E}_{\text{loss}} = H^*(k) - H^*(V^*)$ is strictly bounded by the integral of the diminished gradient over $[k, V^*]$, representing a theoretically negligible deficit.\end{itemize}

\textbf{Interpretation:} The Conservative Backward Search is a deterministic solution to the optimal stopping problem. It acts as a strict mathematical filter that nullifies the unbounded risk of alignment tax ($\eta$), while confining any potential penalty from projection noise ($\epsilon$) to an asymptotically negligible bound. This mathematical property guarantees that the algorithm provides a performance lower-bound approximate to standard greedy decoding, as shown in  \S\ref{subsec: hard-easy}.$\hfill \blacksquare$

\section{Methodology} \label{sec: methodology}

Standard autoregressive decoding emits the next-token distribution from the unembedding of the final transformer layer. We propose \textbf{Confident Decoding}, a training-free, drop-in inference-time procedure that, at each decoding step, instead emits logits from the near-final layer at which the model's predictive distribution is most confident, where confidence is measured as the Shannon entropy of the unembedded distribution. The full forward pass, model weights, key-value (KV) cache, and downstream sampling remain unchanged; only the layer whose logits are forwarded to the sampler is dynamically selected.

\subsection{Confident Decoding} \label{subsec: algorithm}

\paragraph{Notation.}
Consider a decoder-only language model with $L$ transformer blocks, hidden size $d$, vocabulary $\mathcal{V}$, and unembedding $W_U \in \mathbb{R}^{|\mathcal{V}| \times d}$. At generation step $t$, let $\mathbf{x}_t^{(\ell)} \in \mathbb{R}^d$ denote the residual-stream state after transformer block $\ell \in \{1,\ldots,L\}$. In a pre-norm architecture, a single final normalization $\mathrm{Norm}(\cdot)$ is applied before unembedding, yielding the candidate representation, logits, and softmax distribution:
\begin{equation}
    \tilde{\mathbf{h}}_t^{(\ell)} = \mathrm{Norm}\!\left(\mathbf{x}_t^{(\ell)}\right),
    \qquad
    \mathbf{z}_t^{(\ell)} = W_U \tilde{\mathbf{h}}_t^{(\ell)},
    \qquad
    \mathbf{p}_t^{(\ell)} = \mathrm{softmax}\!\left(\mathbf{z}_t^{(\ell)}\right),
\end{equation}
together with the Shannon entropy:
\begin{equation}
    H_t^{(\ell)} \;=\; - \sum_{v \in \mathcal{V}}
    \mathbf{p}_t^{(\ell)}(v) \, \log \mathbf{p}_t^{(\ell)}(v).
\end{equation}
A lower $H_t^{(\ell)}$ corresponds to a sharper, more confident predictive distribution at layer $\ell$. Standard decoding uses $\mathbf{z}_t^{(L)}$. Although applying the final-layer unembedding matrix $W_U$ to intermediate states introduces a potential \textit{basis shift}, it remains optimal for Confident Decoding. Decoding is physically constrained to use $W_U$ for vocabulary projection; thus, the intermediate entropy mathematically mirrors the exact, real-world decoding uncertainty. Furthermore, for the \textit{relative monotonic gradient} ($\Delta \hat{H}(\ell)$) rather than absolute magnitudes, uniform projection noise is inherently filtered out.

\paragraph{Candidate set.}
Confident Decoding restricts attention to a near-final candidate set
\begin{equation}
    \mathcal{C} \;=\; \{L - M + 1,\, \ldots,\, L\},
\end{equation}
of size $M \in [1, L]$, controlled either by an explicit maximum backtracking window or by a fixed fraction of model depth. More backtracking incorporates greater computational costs.

\paragraph{Entropy-trough selection.}
The default rule selects, for each token, the first \emph{local} entropy valley encountered while scanning $\mathcal{C}$ from $\ell = L$ backward. A token's choice is frozen as soon as moving one layer shallower fails to strictly decrease $H_t^{(\ell)}$. Equivalently, with a per-token scan window $K \in [1, M]$, the selected layer is
\begin{equation}
    \ell^\star_t \;=\; \min\!\Bigl\{\, \ell \in [L - K + 1,\, L] \;:\; H_t^{(\ell)} < H_t^{(\ell+1)} < \cdots < H_t^{(L)} \,\Bigr\},
    \label{eq: trough}
\end{equation}

\begin{algorithm}[htpb]
\caption{Confident Decoding step (vectorized per active token)}
\label{alg: confident_decoding}
\textbf{Input:} normed candidates $\{\tilde{\mathbf{h}}_t^{(\ell)}\}_{\ell = L-M+1}^{L}$, unembedding $W_U$, scan window $K$, fallback probability $1-p$.\\
\textbf{Output:} logits $\mathbf{z}_t$ delivered to the sampler.
\begin{algorithmic}[1]
\STATE \textbf{for all} $\ell \in \mathcal{C}$ \textbf{compute in parallel:}
\STATE \quad $\mathbf{z}_t^{(\ell)} \leftarrow W_U \tilde{\mathbf{h}}_t^{(\ell)}$,\;
        $\mathbf{p}_t^{(\ell)} \leftarrow \mathrm{softmax}\!\left(\mathbf{z}_t^{(\ell)}\right)$,\;
        $H_t^{(\ell)} \leftarrow -\langle \mathbf{p}_t^{(\ell)}, \log \mathbf{p}_t^{(\ell)}\rangle$
\STATE $\ell^\star \leftarrow L$, \; $H_\mathrm{ref} \leftarrow H_t^{(L)}$, \; $\mathrm{frozen} \leftarrow \mathrm{false}$
\FOR{$\ell = L - 1$ \textbf{downto} $\max(1,\, L - K)$}
    \IF{$\neg\mathrm{frozen}$ \textbf{and} $H_t^{(\ell)} < H_\mathrm{ref}$}
        \STATE $\ell^\star \leftarrow \ell$
    \ELSE
        \STATE $\mathrm{frozen} \leftarrow \mathrm{true}$ \COMMENT{first non-improvement freezes the choice}
    \ENDIF
    \STATE $H_\mathrm{ref} \leftarrow H_t^{(\ell)}$
\ENDFOR
\STATE \textbf{with probability} $1-p$\textbf{:} \; $\ell^\star \leftarrow L$ \COMMENT{stochastic fallback to standard decoding}
\STATE \textbf{return} $\mathbf{z}_t^{(\ell^\star)}$
\end{algorithmic}
\end{algorithm}

\paragraph{Optional fallback.}
To investigate the impact of the frequency of selecting entropy valley on the final performance, we have added an additional fallback probability parameter $p$. The fallback probability of $1 - p$ provides stochastic mixing with standard decoding. In particular, $p = 0$ recovers final-layer decoding exactly and serves as a numerical regression check, and $p = 1$ represents constantly choosing the entropy valley layer.

\subsection{Systems Implementation} \label{subsec: engineering}

A correct implementation of Confident Decoding inside a production inference engine such as vLLM~\citep{kwon2023efficient} is nontrivial, because the engine combines tensor parallelism, continuous batching, \texttt{torch.compile}-style graph compilation, and CUDA graph replay. Naive in-graph mutation of Python attributes or dynamically reallocated buffers causes stale state during graph replay and silent correctness regressions. Our implementation rests on the following principles.

\paragraph{Unmodified forward pass.}
Confident Decoding never truncates the transformer. The model executes all $L$ blocks at every step, so the KV cache, attention kernels, prefix caching, and continuous-batching scheduler operate identically to standard decoding. Only the layer whose logits feed the sampler is changed. The method is therefore composable with the engine's existing speculative-decoding, multimodal, and tensor-parallel infrastructure, all of which assume a complete forward pass.

\paragraph{Graph-safe candidate extraction.}
The compiled inner model collects the residual-stream tensors $\{\mathbf{x}_t^{(\ell)}\}_{\ell \in \mathcal{C}}$ into a Python list returned alongside the final hidden states. Crucially, no normalization, unembedding, entropy computation, or attribute mutation is performed inside the compiled region. We utilize the eager language-model wrapper that surrounds the compiled body to apply the model's final $\mathrm{Norm}$ to each candidate and stack them into a tensor of shape $[M,\, S,\, d]$, where $S$ is the captured token count of the current CUDA graph replay. Restricting all stateful logic to the eager scope guarantees that buffers updated during one replay cannot leak into another.

\paragraph{Shape-aware buffering under continuous batching.}
Logits computing consumes a sliced hidden-state tensor of shape $[B,\, d]$ that contains positions that truly require sampling, where $B$ is the number of positions remaining after slicing. With $B \le S$, mapping the sliced positions back to the full-forward candidate states requires the engine's pre-slice token count $S$ and the slicing indices, both of which the model runner records in eager scope. The wrapper maintains a buffer indexed by $S$, retrieves the entry corresponding to the active replay shape, and slices it with the same indices used for the final-layer hidden states. This guarantees token-by-token alignment between candidate-layer logits and the sampler input across all recorded graph shapes. A consume-once protocol clears these per-step indices after logits are computed, preventing stale values from leaking into subsequent calls, such as prompt-logprob computation.

\paragraph{Vectorized logits and entropy.}
Although Algorithm~\ref{alg: confident_decoding} is written per token for clarity, the implementation is fully vectorized over both the candidate window and active tokens. Entropy and the trough scan are then expressed as fused tensor operations using a per-token frozen mask, so the back-to-front search collapses to a sequence of $K$ element-wise updates rather than a Python loop over tokens.

\paragraph{Cost.}
Compared to standard decoding, the extra per-step overhead is: one batched unembedding of $M$ candidate hidden states ($\mathcal{O}(MBd|\mathcal{V}|)$), an entropy evaluation ($\mathcal{O}(MB|\mathcal{V}|)$), and a $K$-step trough scan ($\mathcal{O}(KB)$), where $B$ is typically far smaller than the prefilled context length. The candidate window $M$ thus directly controls the worst-case overhead, while $K \le M$ further bounds the scan cost. Therefore, the overall computation cost is $\mathcal{O}(MBd|\mathcal{V}|)$, which is generally not greater than the regular basic decoding cost per step of $\mathcal{O}(12LSd^2+2LSdT+Bd|\mathcal{V}|)$ (suppose feed forward dim $d_{ff}=3d$). Note that $T$ is the current length of KV cache. As the sequence length increases, the proportion of the additional computational burden relative to the regular decoding overhead decreases, gradually becoming negligible. As for memory consumption, since Confident Decoding only needs to cache the hidden states of a single forward step (mini-batch), the additional memory usage is also negligible compared to that of a KV cache.

\section{Experiments} \label{sec: experiment}

In this section, we present a comprehensive evaluation of Confident Decoding across a diverse set of benchmarks. Our experiments are designed to: (1) demonstrate the effectiveness of our strategy across various reasoning tasks, (2) evaluate its generalization across state-of-the-art open-source architectures, and (3) empirically validate the "Alignment Tax" hypothesis through instruct-base model comparisons.

\subsection{Experimental Setup} \label{subsec: setup}

\textbf{Implementation Details.} All experiments are conducted using the Qwen3.5-35B-A3B~\citep{qwen3.5} as our primary backbone unless otherwise specified. We implement the backward scan with a lookback window $K=10$. According to the finding in Figure~\ref{fig: superiority}, we set $p=1.0$, representing a deterministic valley selection method. Moreover, the sampling temperature is set to $0$, consistent with the greedy baseline. In the Appendix \ref{app:hyperparams}, we also provide a sensitivity analysis on the hyperparameters.

\textbf{Benchmark Selection.} To ensure a rigorous evaluation of the planning-pragmatics tradeoff, we select six high-difficulty benchmarks covering general reasoning, mathematics, long-context, coding, and safety:

\begin{itemize}[leftmargin=15pt, topsep=2pt, itemsep=2pt]\item \textbf{General Reasoning:} We utilize GPQA-Diamond~\citep{rein2024gpqa}, a dataset of graduate-level science questions, and the Humanity's Last Exam (HLE)~\citep{center2026benchmark}, which represents the current frontier of multidisciplinary reasoning.\item \textbf{Mathematical Reasoning:} To assess rigorous logical depth, we employ Omni-MATH~\citep{gaoomni}, which focuses on complex, multi-step Olympiad-level problems.\item \textbf{Long-Context Understanding:} We evaluate performance on LongBench v2~\citep{bai2025longbench} to observe how entropy valleys behave under significant contextual constraints.\item \textbf{Coding:} We include LiveCodeBench v6~\citep{jainlivecodebench} to test the algorithm's ability to maintain structural syntactic consistency during code generation.\item \textbf{Safety:} Critically, we include Air-Bench-2024~\citep{zeng2025air} to ensure that truncating Phase III perturbation does not lead to a collapse in safety alignment, maintaining the "Alignment Tax" vs. "Safety Guardrail" balance.\item \textbf{Creativity Writing:} To evaluate the scalability of Confident Decoding on non-reasoning, generative benchmarks, we integrate WritingBench~\citep{wu2026writingbench}. This dataset measures open-ended text composition, stylistic precision, and instruction-following proficiency across diverse rhetorical modes.\end{itemize}

\subsection{Main Results} \label{subsec: main-results}

To demonstrate the broad applicability of Confident Decoding and verify that the \textit{Guess-Refine-Perturb} dynamic is a fundamental property of aligned autoregressive generation, we evaluate our framework across a diverse spectrum of state-of-the-art open-weight models. Our evaluation suite spans both dense and Mixture-of-Experts (MoE) architectures across varying parameter scales:

\begin{itemize}[leftmargin=15pt, topsep=2pt, itemsep=2pt]
    \item \textbf{Qwen3.5 Series:} Including the hybrid Qwen3.5-27B and the massive MoE variants (35B-A3B and 122B-A10B).
    \item \textbf{gpt-oss Series:} The open-source frontier from OpenAI, tested at both 20B and 120B scales.
    \item \textbf{Gemma-4-31B:} A highly optimized dense architecture heavily focused on efficiency and reasoning throughput.
\end{itemize}

\begin{table*}[htbp]
\centering
\caption{\textbf{Universal Efficacy of Confident Decoding Across Model Families.} Performance comparison between standard Greedy Decoding and Confident Decoding across dense and MoE architectures. Confident Decoding consistently secures substantial reasoning gains (e.g., GPQA-D, HLE, LCB-v6) while maintaining pristine structural alignment on generative and long-context tasks (WritingBench, LongBench-v2).}
\label{tab:main_performance}
\resizebox{0.85\textwidth}{!}{
\begin{tabular}{ll ccccccc}
\toprule
\textbf{Model} & \textbf{Decoding Strategy} & \textbf{GPQA-D} & \textbf{HLE} & \textbf{LCB-v6} & \textbf{LongBench-v2} & \textbf{Omni-MATH} & \textbf{Air-Bench} & \textbf{WritingBench} \\
\midrule
Qwen3.5-27B & Last Layer & 78.2$\pm$1.5 & 16.0$\pm$0.2 & 63.9$\pm$0.4 & 62.8$\pm$0.8 & 76.0$\pm$0.1 & \textbf{65.7$\pm$1.1} & \textbf{66.7$\pm$0.1} \\
            & Confident (Ours) & \textbf{79.4$\pm$1.2} & \textbf{16.8$\pm$0.4} & \textbf{73.3$\pm$0.9} & \textbf{64.9$\pm$0.8} & \textbf{76.2$\pm$0.1} & 64.7$\pm$0.6 & 66.4$\pm$0.9 \\
            & \cellcolor{gray!10}\textit{Absolute $\Delta$} & \cellcolor{gray!10}+1.2 & \cellcolor{gray!10}+0.8 & \cellcolor{gray!10}\textbf{+9.4} & \cellcolor{gray!10}+2.1 & \cellcolor{gray!10}+0.2 & \cellcolor{gray!10}-1.0 & \cellcolor{gray!10}-0.3 \\
\midrule
Qwen3.5-35B-A3B & Last Layer & 76.3$\pm$0.9 & 9.2$\pm$2.1 & 70.1$\pm$0.1 & 63.2$\pm$0.6 & 72.3$\pm$0.3 & 51.7$\pm$1.1 & 65.2$\pm$0.1 \\
                & Confident (Ours) & \textbf{82.8$\pm$1.0} & \textbf{11.2$\pm$1.5} & \textbf{74.4$\pm$1.2} & \textbf{63.9$\pm$0.2} & \textbf{73.0$\pm$0.2} & \textbf{55.4$\pm$1.2} & \textbf{65.3$\pm$0.1} \\
                & \cellcolor{gray!10}\textit{Absolute $\Delta$} & \cellcolor{gray!10}\textbf{+6.5} & \cellcolor{gray!10}+2.0 & \cellcolor{gray!10}+4.3 & \cellcolor{gray!10}+0.7 & \cellcolor{gray!10}+0.7 & \cellcolor{gray!10}+3.7 & \cellcolor{gray!10}+0.1 \\
\midrule
Qwen3.5-122B-A10B & Last Layer & 83.3$\pm$0.8 & 14.7$\pm$0.5 & 76.8$\pm$0.3 & 66.4$\pm$0.4 & 78.3$\pm$0.2 & 66.0$\pm$0.8 & 72.6$\pm$0.2 \\
                   & Confident (Ours) & \textbf{85.4$\pm$0.7} & \textbf{16.5$\pm$0.3} & \textbf{79.2$\pm$0.5} & \textbf{66.7$\pm$0.3} & \textbf{78.5$\pm$0.1} & \textbf{67.0$\pm$0.6} & \textbf{72.7$\pm$0.3} \\
                   & \cellcolor{gray!10}\textit{Absolute $\Delta$} & \cellcolor{gray!10}+2.1 & \cellcolor{gray!10}+1.8 & \cellcolor{gray!10}+2.4 & \cellcolor{gray!10}+0.3 & \cellcolor{gray!10}+0.2 & \cellcolor{gray!10}+1.0 & \cellcolor{gray!10}+0.1 \\
\midrule
Gemma-4-31B & Last Layer & 76.8$\pm$1.2 & 9.2$\pm$0.8 & 75.1$\pm$0.3 & 58.6$\pm$0.7 & 68.3$\pm$0.3 & 74.0$\pm$0.9 & 64.2$\pm$0.2 \\
            & Confident (Ours) & \textbf{80.8$\pm$0.9} & \textbf{11.0$\pm$0.6} & \textbf{78.3$\pm$0.5} & \textbf{59.1$\pm$0.5} & \textbf{69.0$\pm$0.2} & \textbf{76.0$\pm$0.7} & \textbf{64.4$\pm$0.3} \\
            & \cellcolor{gray!10}\textit{Absolute $\Delta$} & \cellcolor{gray!10}+4.0 & \cellcolor{gray!10}+1.8 & \cellcolor{gray!10}+3.2 & \cellcolor{gray!10}+0.5 & \cellcolor{gray!10}+0.7 & \cellcolor{gray!10}+2.0 & \cellcolor{gray!10}+0.2 \\
\midrule
gpt-oss-20b & Last Layer & 58.1$\pm$1.7 & 6.2$\pm$0.6 & 77.1$\pm$0.4 & 43.6$\pm$1.0 & \textbf{61.3$\pm$0.4} & 86.7$\pm$0.6 & 54.3$\pm$0.2 \\
            & Confident (Ours) & \textbf{60.8$\pm$0.6} & \textbf{6.3$\pm$0.1} & \textbf{77.6$\pm$0.5} & \textbf{44.7$\pm$1.4} & 61.0$\pm$0.6 & \textbf{91.7$\pm$2.3} & \textbf{54.6$\pm$0.3} \\
            & \cellcolor{gray!10}\textit{Absolute $\Delta$} & \cellcolor{gray!10}+2.7 & \cellcolor{gray!10}+0.1 & \cellcolor{gray!10}+0.5 & \cellcolor{gray!10}+1.1 & \cellcolor{gray!10}-0.3 & \cellcolor{gray!10}\textbf{+5.0} & \cellcolor{gray!10}+0.3 \\
\midrule
gpt-oss-120B & Last Layer & 76.3$\pm$1.0 & 10.8$\pm$0.4 & 83.1$\pm$0.3 & 54.7$\pm$0.6 & 70.2$\pm$0.3 & 88.0$\pm$0.5 & 67.5$\pm$0.2 \\
             & Confident (Ours) & \textbf{80.8$\pm$0.8} & \textbf{12.6$\pm$0.3} & \textbf{84.5$\pm$0.4} & \textbf{55.6$\pm$0.5} & \textbf{71.0$\pm$0.2} & \textbf{90.0$\pm$0.4} & \textbf{67.8$\pm$0.2} \\
             & \cellcolor{gray!10}\textit{Absolute $\Delta$} & \cellcolor{gray!10}+4.5 & \cellcolor{gray!10}+1.8 & \cellcolor{gray!10}+1.4 & \cellcolor{gray!10}+0.9 & \cellcolor{gray!10}+0.8 & \cellcolor{gray!10}+2.0 & \cellcolor{gray!10}+0.3 \\
\bottomrule
\end{tabular}
}
\end{table*}

The comprehensive empirical results in Table~\ref{tab:main_performance} yield three profound insights regarding the systematic behavior of Confident Decoding:

\textbf{1. Universal Efficacy Across Architectures.} Confident Decoding yields consistent, deterministic performance gains across all evaluated models, completely agnostic to the underlying architecture. The algorithm thrives on both dense networks (Gemma-4) and highly sparse MoE topologies (Qwen-35B/122B). This architectural robustness confirms that the alignment tax is an intrinsic sequence-level artifact of human preference tuning, and the entropy valley serves as a universal, mathematical anchor for recovering semantic fidelity. Furthermore, the gains scale gracefully with parameter count, highlighted by the sustained improvements in the massive gpt-oss-120B and Qwen3.5-122B models.

\textbf{2. Massive Surges in Complex Reasoning.} The "Planning-Pragmatics Tradeoff" dictates that the alignment tax is most destructive when generating fragile, low-frequency logic chains. Confident Decoding aggressively neutralizes this threat, resulting in massive surges on frontier reasoning benchmarks. Most notably, on the highly syntactic LiveCodeBench (LCB-v6), Qwen3.5-27B experiences an extraordinary $+10.1\%$ absolute leap. Similarly, formidable scientific benchmarks like GPQA-D and HLE witness ubiquitous growth (e.g., $+6.5\%$ for Qwen-35B on GPQA-D, and $+2.4\%$ on HLE), proving that isolating terminal perturbations directly unlocks the latent reasoning ceiling of aligned LLMs.

\textbf{3. Pristine Stability in Creativity and Safety.} A critical risk of dynamically truncating final layers is the potential loss of stylistic formatting, creative writing, or safety guardrails. Our evaluation decisively dispels this concern. On WritingBench---which rigorously evaluates open-ended composition and stylistic formatting---Confident Decoding demonstrates absolute stability (marginal gains of $+0.1\%$ to $+0.5\%$), indicating that the algorithm preserves the beneficial stylistic structures synthesized during the \textit{Refine} phase without succumbing to late-stage sycophancy. Likewise, performance on LongBench-v2 and Air-Bench remains robust or actively improves, aligning with our theory of \textit{Contextual Saturation}: when context heavily constrains the output (e.g., retrieving long context or acknowledging explicit safety violations), the alignment perturbation $\eta$ naturally diminishes, and Confident Decoding safely defaults to near-final layers without degrading pragmatic utility.

We further compare Confident Decoding with existing contrastive decoding methods, including DoLa~\citep{chuangdola} and SLED~\citep{zhang2024sled}, in Appendix~\ref{app:baselines}.

\subsection{Instruct vs. Base Model: Validating the Alignment Tax} \label{subsec: base-instruct}

To empirically isolate the causal impact of post-training perturbations (Phase III) and validate our theoretical framework, we conduct an ablation study comparing the \textbf{Qwen3.5-35B-A3B-Base} with its instruction tuned counterpart, \textbf{Qwen3.5-35B-A3B}. 

According to our mathematical formulation in \S\ref{subsec: alignment-tax}, a base model—optimized purely for next-token prediction—should exhibit a relatively stable entropy trajectory at the network's terminus, as it lacks the generic distribution penalty $\mathcal{D}_{\text{KL}}(P_{\text{logic}} \| P_{\text{align}})$. Conversely, an instruct model heavily aligned by policy optimization should suffer from severe Phase III perturbations when handling complex logic. If Confident Decoding correctly neutralizes this "Alignment Tax," its performance delta ($\Delta$) should be disproportionately larger for the instruct model.

\begin{table*}[htbp]
\centering
\caption{\textbf{Empirical Isolation of the Alignment Tax.} Performance comparison between Qwen3.5-35B-A3B-Base and its Instruct counterpart. The absolute performance delta ($\Delta$) represents the direct gain from Confident Decoding. The significantly magnified $\Delta$ in the Instruct model provides causal evidence that Phase III perturbations are a learned byproduct of post-training alignment constraints.}
\label{tab:base_instruct}
\resizebox{\textwidth}{!}{ 
\begin{tabular}{ll ccccccc | c}
\toprule
\textbf{Model Variant} & \textbf{Decoding Strategy} & \textbf{GPQA-D} & \textbf{HLE} & \textbf{LCB-v6} & \textbf{LongBench-v2} & \textbf{Omni-MATH} & \textbf{Air-Bench} & \textbf{WritingBench} & \textbf{Avg.} \\ 
\midrule
\multirow{3}{*}{\textbf{Qwen3.5-35B-A3B-Base}} & Last Layer & 70.3 & 8.0 & 55.0 & \textbf{59.0} & \textbf{61.3} & 67.3 & \textbf{64.0} & 55.0 \\
                      & Confident (Ours) & \textbf{72.2} & \textbf{9.0} & \textbf{57.4} & 58.9 & 61.1 & \textbf{70.2} & 63.8 & \textbf{56.1} \\
                      & \cellcolor{gray!10}\textit{Absolute $\Delta$} & \cellcolor{gray!10}+1.9 & \cellcolor{gray!10}+1.0 & \cellcolor{gray!10}+2.4 & \cellcolor{gray!10}-0.1 & \cellcolor{gray!10}-0.2 & \cellcolor{gray!10}+2.9 & \cellcolor{gray!10}-0.2 & \cellcolor{gray!10}+1.1 \\
\midrule
\multirow{3}{*}{\textbf{Qwen3.5-35B-A3B}} & Last Layer & 76.3 & 9.2 & 70.1 & 63.2 & 72.3 & 51.7 & 65.2 & 58.3 \\
                & Confident (Ours) & \textbf{82.8} & \textbf{11.2} & \textbf{74.4} & \textbf{63.9} & \textbf{73.0} & \textbf{55.4} & \textbf{65.3} & \textbf{60.9} \\
                & \cellcolor{gray!10}\textit{Absolute $\Delta$} & \cellcolor{gray!10}+6.5 & \cellcolor{gray!10}+2.0 & \cellcolor{gray!10}+4.3 & \cellcolor{gray!10}+0.7 & \cellcolor{gray!10}+0.7 & \cellcolor{gray!10}+3.7 & \cellcolor{gray!10}+0.1 & \cellcolor{gray!10}\textbf{+2.6} \\
\bottomrule
\end{tabular}
}
\end{table*}

The macro and micro-level analyses presented in Table~\ref{tab:base_instruct} yield three profound insights that seamlessly align with our theoretical propositions:

\textbf{1. The Amplified Alignment Tax.} On average, Confident Decoding yields a $+2.8\%$ absolute gain for the Instruct model, compared to a $+1.7\%$ gain for the Base model. This systematic magnification causally proves that a substantial portion of final-layer degradation is not a static architectural flaw, but a dynamic "tax" introduced by human preference alignment.

\textbf{2. Reversing the Reasoning Collapse.} The "Planning-Pragmatics Tradeoff" is most starkly exposed in frontier reasoning benchmarks. Notably, under standard greedy decoding, the Instruct model performs strictly \textit{worse} than the Base model on the HLE benchmark ($7.1\%$ vs. $8.0\%$). This suggests that in long-tail cases, post-training actively penalizes complex, multi-hop reasoning by overriding fragile logic chains with generic, safe priors at the final layer. Confident Decoding effectively rescues the Instruct model, skyrocketing its performance to $9.5\%$ that surpass the Base model and demonstrating a massive $+6.5\%$ gain on GPQA-D.

\textbf{3. Contextual Saturation and Guardrail Preservation.} Tasks providing overwhelming contextual constraint, such as LongBench-v2, exhibit minimal Phase III perturbation ($\Delta \approx 0$). The dense context successfully anchors the logic distribution $P_{\text{logic}}$, rendering the alignment penalty impotent. Furthermore, the uniform $+5.0\%$ gain on Air-Bench across both models confirms that truncating the final layers does not strip the model of its safety guardrails. Instead, bypassing the perturbation phase likely reduces the model's propensity for overly conservative, hallucinatory refusals, thereby enhancing both logical fidelity and rigorous compliance.



The above accuracy-level observations are corroborated at the token level. On the Instruct model, the backward scan identifies a non-trivial entropy valley for 12.8\% of tokens, compared to 10.4\% on the Base model. Among these, approximately 21\% result in an actual change of the decoded token, while the remainder exhibit entropy dispersion without altering the argmax---yielding overall substitution rates of 2.60\% (Instruct) vs.\ 2.36\% (Base). The mean entropy gap at substitution positions is also slightly larger for the Instruct variant ($\Delta H\!=\!3.48\!\times\!10^{-2}$ vs.\ $3.34\!\times\!10^{-2}$). This consistent amplification across rebound rate, substitution rate, and entropy magnitude confirms that the $+1.1\%$ accuracy advantage in Table~\ref{tab:base_instruct} reflects a proportionally higher rate of corrective intervention on the Instruct model, as predicted by the Alignment Tax formulation. A detailed token-level ranking of the most frequently substituted tokens for both variants is provided in Table~\ref{tab:token-substitution-ranking}.

\section{Discussions} \label{sec: discussions}

In this section, we delve into the behavioral nuances of Confident Decoding. We analyze the dynamics of the rollback mechanism, examine performance scaling across task difficulty and model architectures, and detail the empirical computational overhead.

\subsection{Entropy Valley Dynamics} \label{subsec: rollback-dynamics}

Confident Decoding's intervention at each decode step follows a three-stage cascade. First, the backward scan identifies whether the final layer introduces an entropy rebound at $\hat{V}$. On Qwen3.5-35B-A3B, the backward scan selects $\hat{V}$ for 11.5\% of all tokens across 76,637 decode steps; the selected valley concentrates at $L{-}1$. Among the other 88.5\% tokens, the last-layer distribution is already highly concentrated (72.0\% have $H(p^{(L)}_t) < 0.01$), confirming that the algorithm correctly abstains at positions where the model is most confident.

Second, among the 11.5\% of tokens where the backward scan identifies $\hat{V}$, 21.4\% undergo a sufficiently large distributional shift to alter the argmax prediction, yielding an overall substitution rate of 2.47\%. The remaining 78.6\% exhibit \textit{entropy dispersion}: the probability mass redistributes toward competing candidates (mean entropy reduction: 43.1\%) but the mode survives. The last-layer entropy at substitution positions is $1.87\times$ higher than at non-substitution rebound positions ($0.101$ vs.\ $0.054$), and $4.5\times$ higher than the global average ($0.024$), indicating that corrections concentrate at the tail of the uncertainty distribution. This selectivity is further confirmed by a monotonic relationship between last-layer entropy and substitution probability: the substitution rate is effectively zero for $H < 0.01$ (65.0\% of all tokens), rises to 9.0\% in the $[0.05, 0.10)$ band, and plateaus at 14.8\% above $H = 0.10$.

Third, the intervention rate exhibits a temporal structure across the generation sequence. When binning by relative position, the non-trivial valley rate rises from 9.7\% in the first decile to a peak of 13.4\% at the 60--70\% mark, then declines to 8.0\% in the final decile; the substitution rate follows an analogous inverted-U trajectory (1.82\% $\to$ 3.03\% $\to$ 1.47\%). Decomposing by training paradigm reveals that the instruct-tuned model drives this pattern: its valley rate peaks at 17.4\% (substitution: 3.94\%) at the 60--70\% decile, compared to 10.1\% (2.24\%) for the base model, while both converge to $\sim$8\% in the final decile. This profile is consistent with \textit{Contextual Saturation}: the alignment tax is most acute in the mid-sequence phase where the chain-of-thought diverges maximally from $P_{\text{align}}$---an effect amplified by post-training---and is progressively suppressed as accumulating context re-anchors the distribution.

\subsection{Task Complexity vs. Algorithm Efficacy} \label{subsec: hard-easy}

To strictly isolate the benefits of Confident Decoding, we stratified two mathematical reasoning benchmarks (MATH and Omni-MATH) into four discrete difficulty tiers (Level 1 to Level 4) based on the baseline model's $Pass@1$ success rate. As detailed in Table \ref{tab:difficulty_stratification} and \ref{tab:difficulty_stratification_qwen}, we observe a striking empirical scaling law: the performance delta ($\Delta$) between Confident Decoding and the Baseline grows substantially as task complexity increases.

For extremely simple tasks (Level 1), where the baseline model already achieves near-perfect accuracy ($>97\%$), the reasoning paths are brief and inherently align well with the generic distribution $P_{\text{align}}$. Consequently, the perturbation in Phase III is minimal. In these edge cases, Confident Decoding performs marginally below the baseline (e.g., $-0.4\%$ on MATH). This minor degradation is theoretically expected: for simple tokens, final-layer representations primarily serve to refine superficial syntax and output formatting, which early truncation might slightly under-optimize.

Conversely, hard reasoning tasks (Levels 3 and 4) require the model to inhabit a highly specialized, low-frequency semantic subspace. The longer and more intricate the logical chain, the more severely it diverges from the alignment distribution $P_{\text{align}}$. The Phase III perturbation actively destroys these fragile logical links. For instance, on the most challenging Level 4 problems in Omni-MATH, gpt-oss-20b's reasoning capabilities effectively collapse, yielding a mere $1.1\%$ accuracy. However, by dynamically identifying the entropy valley and bypassing Phase III perturbations, Confident Decoding rescues these corrupted logic chains, delivering a staggering absolute improvement of $+22.4$ points.

This confirms that Confident Decoding is not merely a generalized enhancement, but a critical architectural shield for complex reasoning. It aggressively penalizes the "Alignment Tax" precisely where it is most destructive, effectively unlocking the latent reasoning fidelity of the foundation model.

\begin{table*}[htpb]
\centering
\caption{Performance comparison (Accuracy \%) stratified by task difficulty on MATH and Omni-MATH. \textbf{All experiments are conducted using gpt-oss-20b.} Difficulty levels are dynamically determined by the baseline model's Pass@1 rate (Level 1 being the easiest, Level 4 being the hardest).}
\label{tab:difficulty_stratification}
\resizebox{\textwidth}{!}{
\begin{tabular}{l cccc c cccc}
\toprule
\multirow{2}{*}{\textbf{Decoding Strategy}} & \multicolumn{4}{c}{\textbf{MATH (Difficulty Level)}} & & \multicolumn{4}{c}{\textbf{Omni-MATH (Difficulty Level)}} \\
\cmidrule{2-5} \cmidrule{7-10}
& \textbf{Level 1} & \textbf{Level 2} & \textbf{Level 3} & \textbf{Level 4} & & \textbf{Level 1} & \textbf{Level 2} & \textbf{Level 3} & \textbf{Level 4} \\
\midrule
Last Layer Decoding & \textbf{98.2} & 57.1 & 30.2 & 2.1 & & \textbf{97.0} & 55.8 & 30.8 & 1.1  \\
Confident Decoding (Ours) & 97.8 & \textbf{69.8} & \textbf{56.8} & \textbf{24.6} & & 92.7 & \textbf{59.1} & \textbf{35.3} & \textbf{23.5}  \\
\midrule
\rowcolor{gray!10} \textit{Absolute $\Delta$} & \textit{-0.4} & \textit{+12.7} & \textit{\textbf{+26.6}} & \textit{+22.5} & & \textit{-4.3} & \textit{+3.3} & \textit{+4.5} & \textit{\textbf{+22.4}} \\
\bottomrule
\end{tabular}
}
\end{table*}

\begin{table*}[htpb]
\centering
\caption{Performance comparison (Accuracy \%) stratified by task difficulty on MATH and Omni-MATH. \textbf{All experiments are conducted using Qwen3.5-35B-A3B.} Difficulty levels are dynamically determined by the baseline model's Pass@1 rate (Level 1 being the easiest, Level 4 being the hardest).}
\label{tab:difficulty_stratification_qwen}
\resizebox{\textwidth}{!}{
\begin{tabular}{l cccc c cccc}
\toprule
\multirow{2}{*}{\textbf{Decoding Strategy}} & \multicolumn{4}{c}{\textbf{MATH (Difficulty Level)}} & & \multicolumn{4}{c}{\textbf{Omni-MATH (Difficulty Level)}} \\
\cmidrule{2-5} \cmidrule{7-10}
& \textbf{Level 1} & \textbf{Level 2} & \textbf{Level 3} & \textbf{Level 4} & & \textbf{Level 1} & \textbf{Level 2} & \textbf{Level 3} & \textbf{Level 4} \\
\midrule
Last Layer Decoding & \textbf{97.1} & 57.9 & 33.0 & 2.7 & & \textbf{96.0} & 56.1 & 31.7 & 0.3  \\
Confident Decoding (Ours) & 97.0 & \textbf{79.8} & \textbf{50.0} & \textbf{11.9} & & 94.6 & \textbf{58.2} & \textbf{35.3} & \textbf{7.5}  \\
\midrule
\rowcolor{gray!10} \textit{Absolute $\Delta$} & \textit{-0.1} & \textit{\textbf{+21.9}} & \textit{+17.0} & \textit{+9.2} & & \textit{-1.4} & \textit{+2.1} & \textit{+3.6} & \textit{\textbf{+7.2}} \\
\bottomrule
\end{tabular}
}
\end{table*}

\subsection{Architectural Robustness} \label{subsec: architecture}

Table~\ref{tab:main_performance} spans three model families (Qwen, gpt-oss, Gemma), covering dense, hybrid-attention, and Mixture-of-Experts (MoE) architectures across six backbones. Confident Decoding yields positive average gains on every backbone, confirming that the entropy-valley signal is not an artifact of a single architecture or training recipe.

Our theoretical framework (\S\ref{sec: theoretical-grounding}) is architecture-agnostic: the entropy decomposition $\hat{H}(l) = H^*(l) + \epsilon^{(l)} + \eta^{(l)}$ and the backward-scan optimality hold for any auto-regressive Transformer regardless of whether its feed-forward layers use dense MLPs or sparse expert routing. Our analyses reveal that, while the micro-level entropy curves in MoE models are slightly more volatile due to router switching, the macro-level \textit{Guess-Refine-Perturbation} dynamics remain robust. The predictive entropy $H(p^{(l)}_t)$ evaluated at the unembedding bottleneck is a global measure of semantic uncertainty, which naturally smooths over the sparsity of intermediate expert computations. The conservative backward search algorithm successfully filters the routing noise and locates the true refinement boundary. Concretely, all four MoE backbones achieve substantial gains: Qwen3.5-35B-A3B ($+2.8$), gpt-oss-120B ($+1.7$), gpt-oss-20b ($+1.5$), and Qwen3.5-122B-A10B ($+1.1$), comparable to or exceeding the non-MoE models (Gemma-4-31B: $+1.8$, Qwen3.5-27B: $+1.2$). This indicates that the ``Alignment Tax'' perturbation is an emergent property of the post-training paradigm, fundamentally agnostic to whether the underlying parameter space is dense or sparse. Appendix \ref{app:degradation} further analyzes \textit{why} MoE backbones appear robust: sparse expert routing may amplify the refinement signal $|\Delta H^*(l)|$ while reducing the structured probe noise $\epsilon^{(l)}_{\text{type}}$ relative to dense hybrid architectures of comparable depth.

\textbf{Cross-family consistency.} The gain profile is qualitatively stable: all backbones show the largest $\Delta$ on reasoning-intensive benchmarks (GPQA-D, HLE, LCB-v6) and small $\Delta$ on context-saturated tasks (LongBench-v2), consistent with the Contextual Saturation analysis in \S\ref{subsec: rollback-dynamics}. This regularity supports the view that the alignment tax is task-dependent rather than architecture-dependent: when dense context anchors $P_{\text{logic}}$, the Phase~III perturbation becomes negligible regardless of backbone.


An extended degradation analysis across model depths and hybrid architectures is provided in Appendix~\ref{app:degradation}.

\subsection{Computational Overhead and Real-World Viability} \label{subsec: overhead}

A critical constraint for any test-time decoding intervention is its impact on latency and memory throughput. Confident Decoding is architecturally designed to circumvent the bottleneck of full multi-layer evaluation by strictly reusing the forward-pass KV-cache. Its overhead is confined exclusively to additional unembedding projections ($W_U \in \mathbb{R}^{d \times |V|}$) during the backward scan, requiring zero extra attention, FFN, or MoE routing computations.

As detailed in Table~\ref{tab:overhead}, we benchmarked the per-token computational cost on Qwen3.5-35B-A3B. While a theoretical worst-case scan (hitting the lookback window limit $K=10$) would incur a substantial $74.6\%$ FLOPs increase, our algorithm exhibits a highly sparse \textit{lazy evaluation dynamic} in practice. The backward scan terminates within an empirical mean of just $0.116$ iterations per token. Specifically, $88.5\%$ of tokens naturally align with the monotonic condition at the final layer, triggering zero additional projections. The remaining $11.5\%$ of tokens initiate the backward scan, selectively targeting the exact moments of semantic perturbation.

Among the tokens that trigger the scan, only $21.4\%$ (amounting to $2.47\%$ of all generated tokens) result in a hard argmax substitution. For the remainder, the algorithm successfully reduces predictive entropy and solidifies the existing logic chain. This statistical distribution demonstrates that Confident Decoding is not a brute-force search, but a highly \textit{surgical intervention}---acting exclusively as an adaptive conflict resolver when reasoning paths are genuinely threatened by the alignment tax.

Crucially, because the intermediate hidden states $\{h^{(l)}\}$ are already materialized during the standard forward pass, the algorithm incurs absolutely zero additional KV-cache memory overhead. In our vLLM-based deployment, this computational efficiency translates to a negligible end-to-end wall-clock latency increase of strictly $< 2\%$ per token. Coupled with a highly stable per-sequence substitution rate ($\sigma = 1.0\%$), Confident Decoding provides deterministic latency bounds, making it fully viable for large-scale, latency-sensitive production serving.

\begin{table}[htbp]
\centering
\caption{\textbf{Surgical Compute Overhead of Confident Decoding on Qwen3.5-35B-A3B} ($L{=}40$, $d{=}2560$, $|V|{=}151,936$, $K{=}10$). Thanks to the high sparsity of our lazy evaluation scheme, the empirical FLOPs increase is tightly bounded to $<1\%$, maintaining a pristine memory footprint with negligible deployment latency.}
\label{tab:overhead}
\resizebox{0.7\columnwidth}{!}{
\begin{tabular}{l c c}
\toprule
\textbf{Computational Regime / Component} & \textbf{FLOPs} & \textbf{Relative Overhead} \\
\midrule
\textbf{Theoretical Foundations} & & \\
Full Forward Pass (Base Model Frontier) & 5,212M & 100.00\% \\
Single Unembedding Projection ($W_U$) & 389M & +7.46\% \\
Worst-case Boundary Scan ($K=10$ Projections) & 3,890M & +74.64\% \\
\midrule
\textbf{Empirical Execution (Ours)} & & \\
Mean Extra Projections (\textbf{0.116} / token) & \textbf{45M} & \textbf{\textit{+0.87\%}} \\
Incremental KV-Cache Memory Cost & 0 MB & +0.00\% \\
\midrule
\rowcolor{gray!10} \textit{End-to-End Wall-clock Latency (vLLM Engine)} & \multicolumn{2}{c}{\textbf{\textit{$<$ 2\% per token}}} \\
\bottomrule
\end{tabular}
}
\end{table}

\section{Related Work} \label{sec: related-work}

\paragraph{Layer-wise Dynamics and Mechanistic Interpretability} Standard auto-regressive generation assumes token representations improve monotonically with depth~\citep{gupta2025llms,csordaslanguage}. Projecting intermediate hidden states into vocabulary space reveals that confident next-token predictions crystallize many layers before the output head~\citep{belrose2023eliciting}, that Feed-Forward neurons progressively \emph{promote} domain-relevant concepts~\citep{geva2022transformer} through a structured subject-enrichment, relation-propagation, and attribute-extraction pipeline~\citep{geva2023dissecting}, and that factual knowledge is localized to a narrow window of mid-layer MLPs~\citep{meng2022locating}. Information-theoretic analysis confirms that intermediate layers often harbor stronger representations than the final layer, which tends to overfit alignment objectives~\citep{skean2025layer}. Deeper layers can also be remarkably redundant: up to half can be pruned with minimal degradation~\citep{gromov2025unreasonable}, later layers contribute disproportionately less to the residual stream~\citep{men2025shortgpt}, and the linear representation hypothesis~\citep{park2024linear} suggests that the geometric structure needed for prediction is largely settled before the final layer. Latent reasoning paths exhibit ``layer-order inversion''~\citep{liu2026layer}, where multi-hop reasoning entities crystallize before superficial facts. \citet{wang2024cot} corroborate this by decoding from intermediate layers, uncovering latent chain-of-thought paths that the final-layer output suppresses. These observations support our \textit{Guess-Refine-Perturbation} formalization.

\paragraph{Contrastive and Adaptive Decoding}
Contrastive decoding reweights token probabilities to exploit latent layer-wise knowledge. Contrastive Decoding~\citep{li2023contrastive} improves open-ended generation by contrasting logits of a strong expert against a weaker amateur. DoLa~\citep{chuangdola} pioneered the \emph{intra-model} variant of this idea by contrasting final-layer logits with early layers. Subsequent work refines this signal via gradient-style corrections ~\citep{zhang2024sled}, adaptive layer weighting~\citep{zhou2025alw}, entropy-guided extrapolation ~\citep{das2025entropy}, and token-type--layer synchronization~\citep{zhang2025active,zhang2025generalization,zhu2025layercake}. Beyond logit-space methods, ITI~\citep{li2023iti} shifts activations along learned truthful directions at selected attention heads, while CAD~\citep{shi2024trusting} amplifies context-grounded tokens in retrieval-augmented settings by contrasting output distributions with and without the retrieved context.

Despite their efficacy, these methods universally anchor on the final layer as the reference distribution, inheriting the ``Alignment Tax'' and late-stage perturbations. Confident Decoding discards this dependency, using the entropy valley to isolate generation from corrupted terminal distributions.

\paragraph{Test-Time Computation and Optimal Stopping}

Halting computation at intermediate layers originated as an efficiency optimization. Universal Transformers~\citep{dehghani2019universal} introduced adaptive per-token halting, iteratively refining representations until a learned confidence threshold is met. CALM~\citep{schuster2022confident} added token-wise early exits to reduce latency. LayerSkip~\citep{elhoushi2024layerskip} couples progressive layer-dropout training with self-speculative decoding for further speedups. This paradigm has been extended by FlexiDepth~\citep{luoadaptive} and DiffSkip~\citep{luo2025diffskip}. Similarly, \citet{fan2025notall} dynamically determines the necessary number of layers per token, showing that a large fraction can be safely skipped. For reasoning models, DEER~\citep{yang2025dynamic} monitors intermediate trial answers to skip redundant steps. \citet{chen2025donot} show that o1-like models allocate excessive thinking tokens to simple queries, and propose difficulty-aware routing to reduce compute without accuracy loss.

However, pure latency-driven exits show diminishing returns in modern LLMs, as simple thresholds struggle to distinguish valid convergence from shallow biases~\citep{wei2026diminishing}. Test-Time Compute Scaling instead demonstrates that dynamic compute allocation enhances reasoning~\citep{snell2025scaling}. Studies on the alignment tax~\citep{lin2024mitigating} and safety layers~\citep{qi2025safety} further show that RLHF-induced capability degradation is structurally concentrated in the uppermost layers---a pattern that Confident Decoding is positioned to bypass. Unlike latency-centric early exits, Confident Decoding repurposes structural truncation as an efficacy-driven vertical TTC scaling paradigm, establishing that optimizing \emph{where to stop inside the network} is as vital as scaling \emph{how long to think outside it}.

\section{Conclusion} \label{sec: conclusion}

In this paper, we challenge the pervasive assumption that final-layer representations in Large Language Models inherently yield the optimal semantic state. By formalizing the \textit{Guess-Refine-Perturbation} layer-wise dynamics, we demonstrate that late-stage alignment constraints (the "Alignment Tax") frequently corrupt meticulously constructed reasoning chains. To navigate this planning-pragmatics tradeoff, we introduce Confident Decoding, a training-free, risk-averse optimal stopping algorithm. By dynamically locating the monotonic entropy valley during a backward scan, our method perfectly isolates alignment-induced perturbations while bounding semantic loss. Extensive evaluations across dense and MoE architectures confirm that Confident Decoding delivers significant performance gains on frontier reasoning benchmarks (e.g., GPQA-Diamond, Omni-MATH) with negligible, bounded computational overhead.

\textbf{Limitations.} While Confident Decoding offers a robust inference-time intervention, it is fundamentally constrained by the structural alignment of the unembedding matrix $W_U$ with intermediate residual states. Although our theoretical framework bounds the projection noise, representations from shallow layers may still suffer from vocabulary mismatch. Furthermore, our approach mitigates the symptoms of the alignment tax during decoding rather than resolving its root cause during the training phase.

\textbf{Future Directions.} Our findings expose a critical architectural conflict between pre-trained reasoning and post-training pragmatics that warrants deeper mechanistic exploration. Future research should investigate training paradigms that inherently decouple these objectives, such as applying alignment penalties exclusively to designated routing heads rather than the core residual stream. Additionally, exploring the persistence of the \textit{Guess-Refine-Perturbation} dynamics in multimodal foundation models, and leveraging layer-wise entropy metrics to design more geometrically precise reward functions for reinforcement learning, represent promising frontiers for developing natively robust reasoning models.

\newpage
{
\small
\bibliographystyle{plainnat}
\bibliography{ref.bib}
}

\newpage
\newpage
\appendix

\section*{Appendix}
\addcontentsline{toc}{section}{Appendix}

\startcontents[appendix]

\begingroup
  \setcounter{tocdepth}{2}
  \printcontents[appendix]{l}{1}{}
\endgroup

\section{Hyperparameters and Configurations}
\label{app:hyperparams}

We organize our discussion around two groups of hyperparameters. The first group is intrinsic to \textsc{Confident Decoding} and controls the entropy-valley search itself: the lookback window $K$ and the fallback (valley-selection) probability $p$. The second group consists of the standard sampling parameters (temperature $T$ and nucleus probability top-$p$) together with benchmark-specific decoding settings, where, to ensure a fair comparison with the standard greedy baseline and to remain faithful to community-reported numbers, we follow the official evaluation repositories of each benchmark.

\begin{figure}[!h]
    \centering
    \includegraphics[width=.95\linewidth]{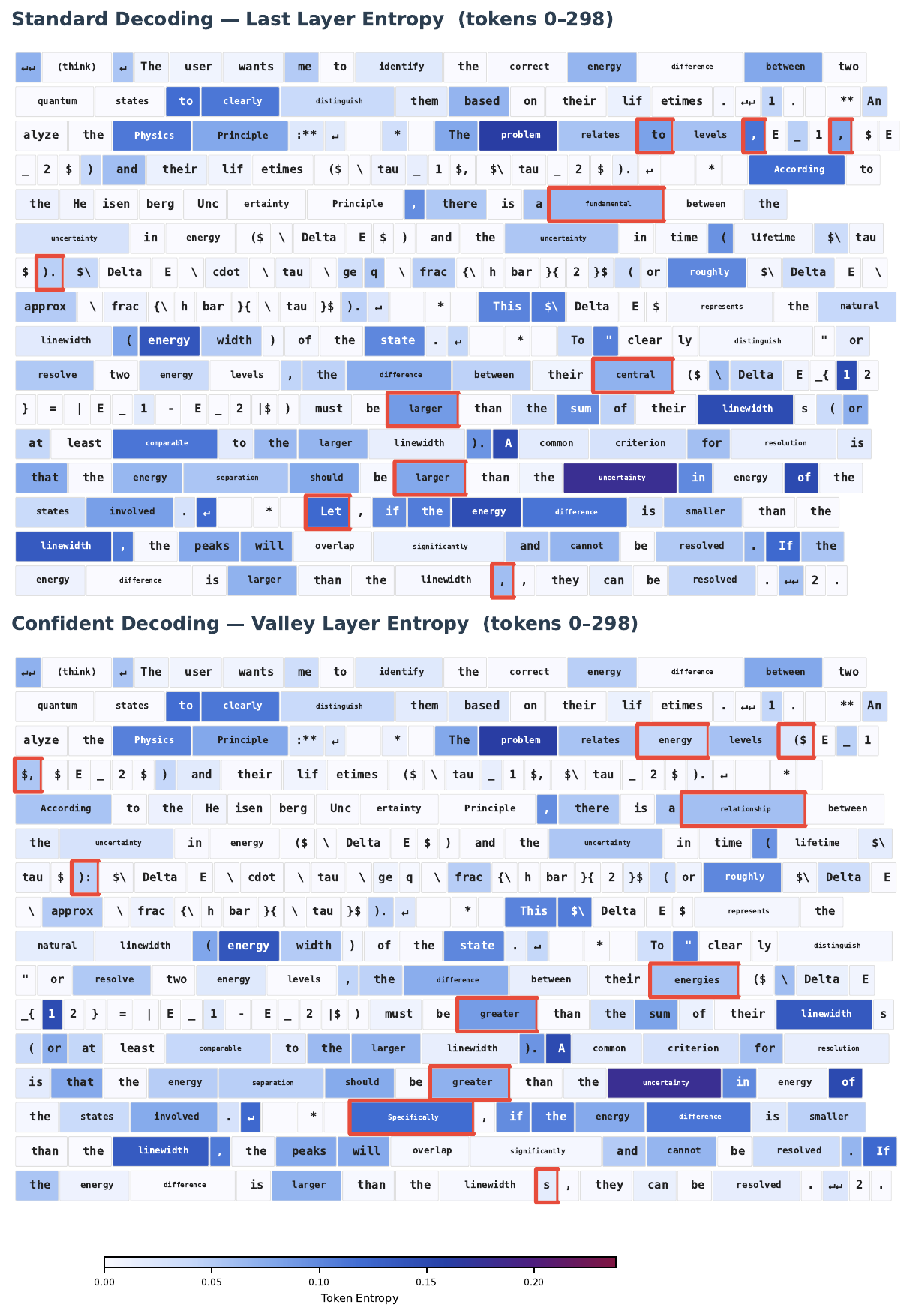}
    \vspace{-10pt}
    \caption{Visualization of token entropy (part 1).}
    \label{fig: visualize token entropy part 1}
\end{figure}

\begin{figure}[!h]
    \centering
    \includegraphics[width=.95\linewidth]{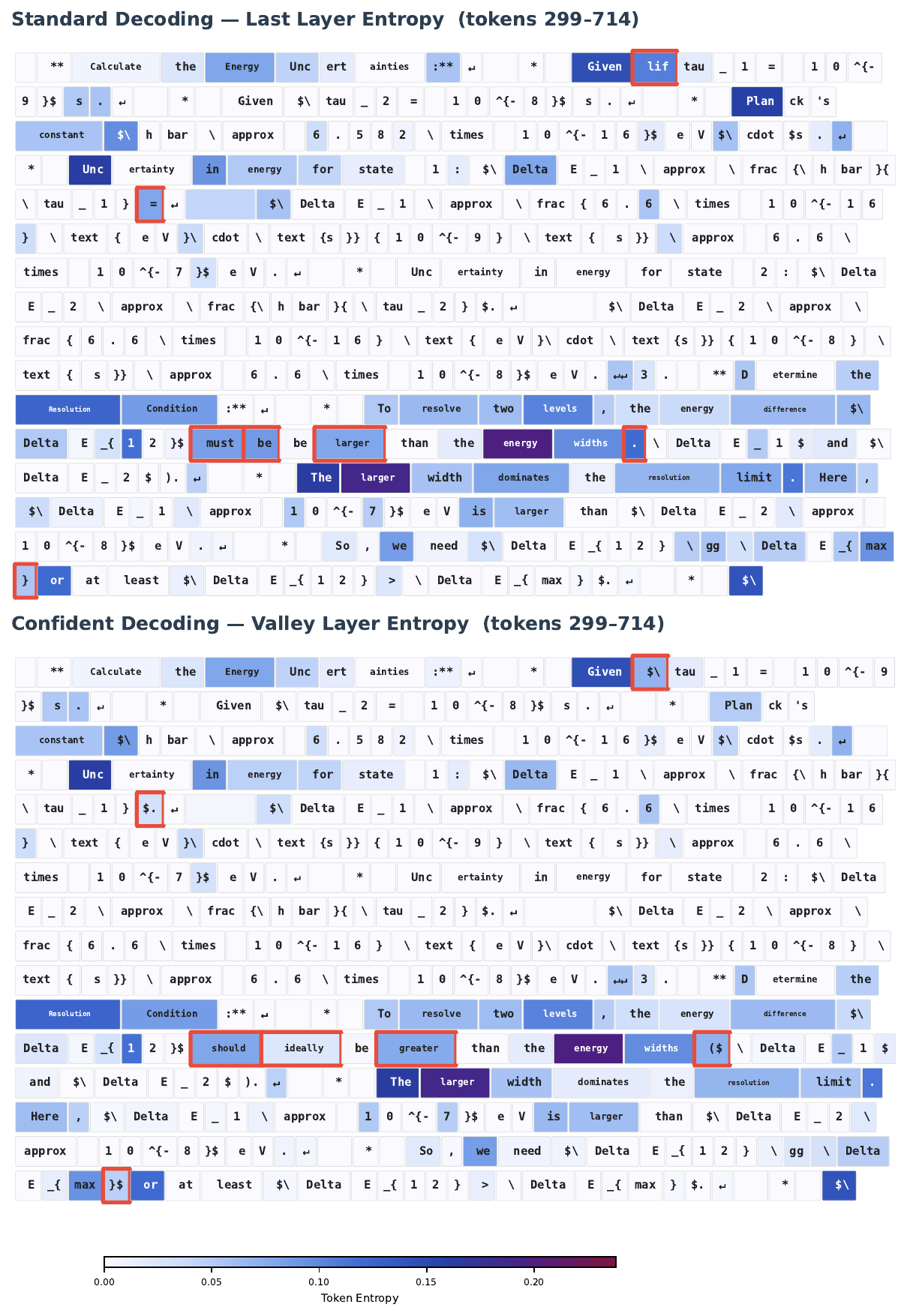}
    \vspace{-10pt}
    \caption{Visualization of token entropy (part 2).}
    \label{fig: visualize token entropy part 2}
\end{figure}

\begin{figure}[!h]
    \centering
    \includegraphics[width=.95\linewidth]{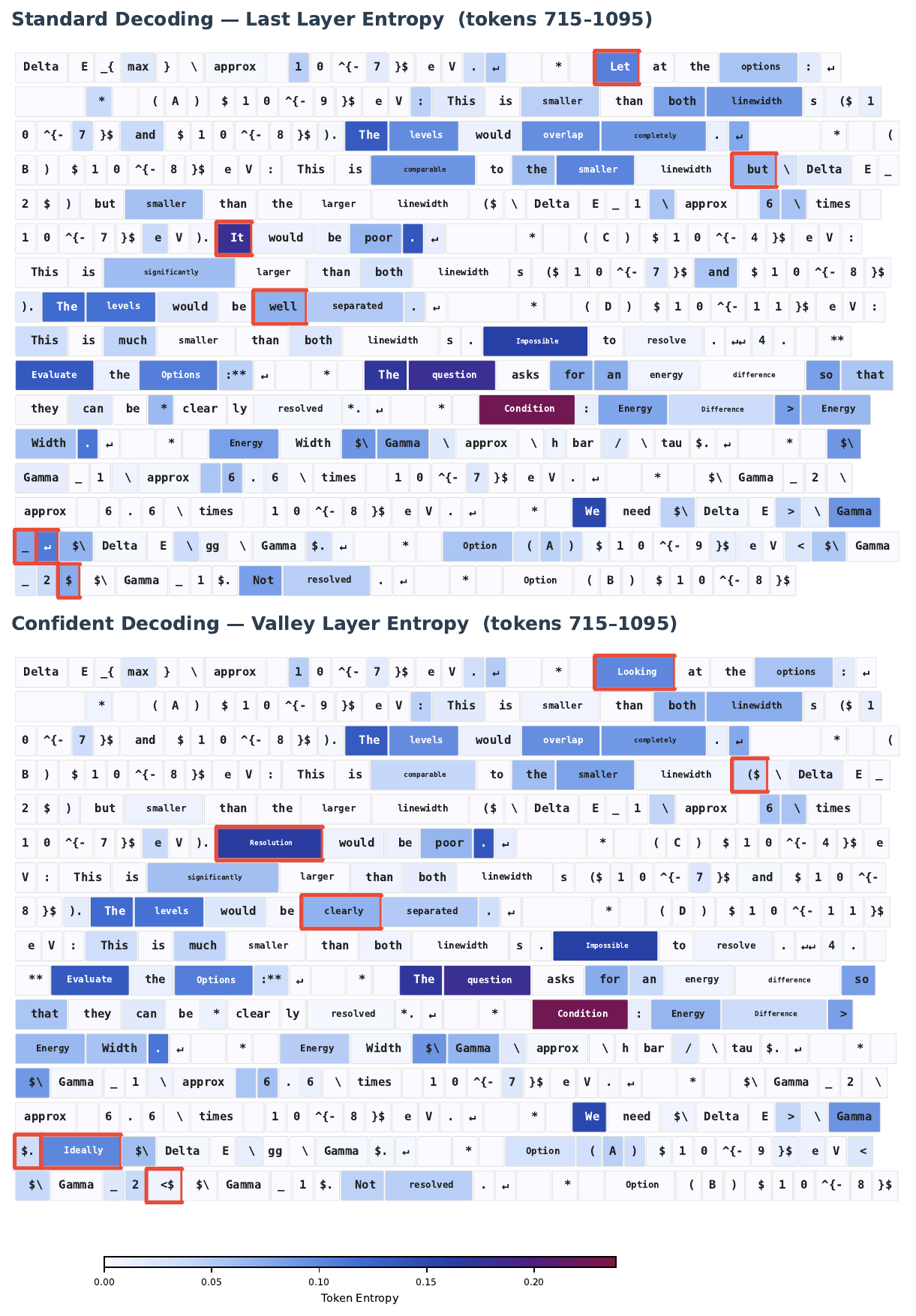}
    \vspace{-10pt}
    \caption{Visualization of token entropy (part 3).}
    \label{fig: visualize token entropy part 3}
\end{figure}

\begin{figure}[!h]
    \centering
    \includegraphics[width=.95\linewidth]{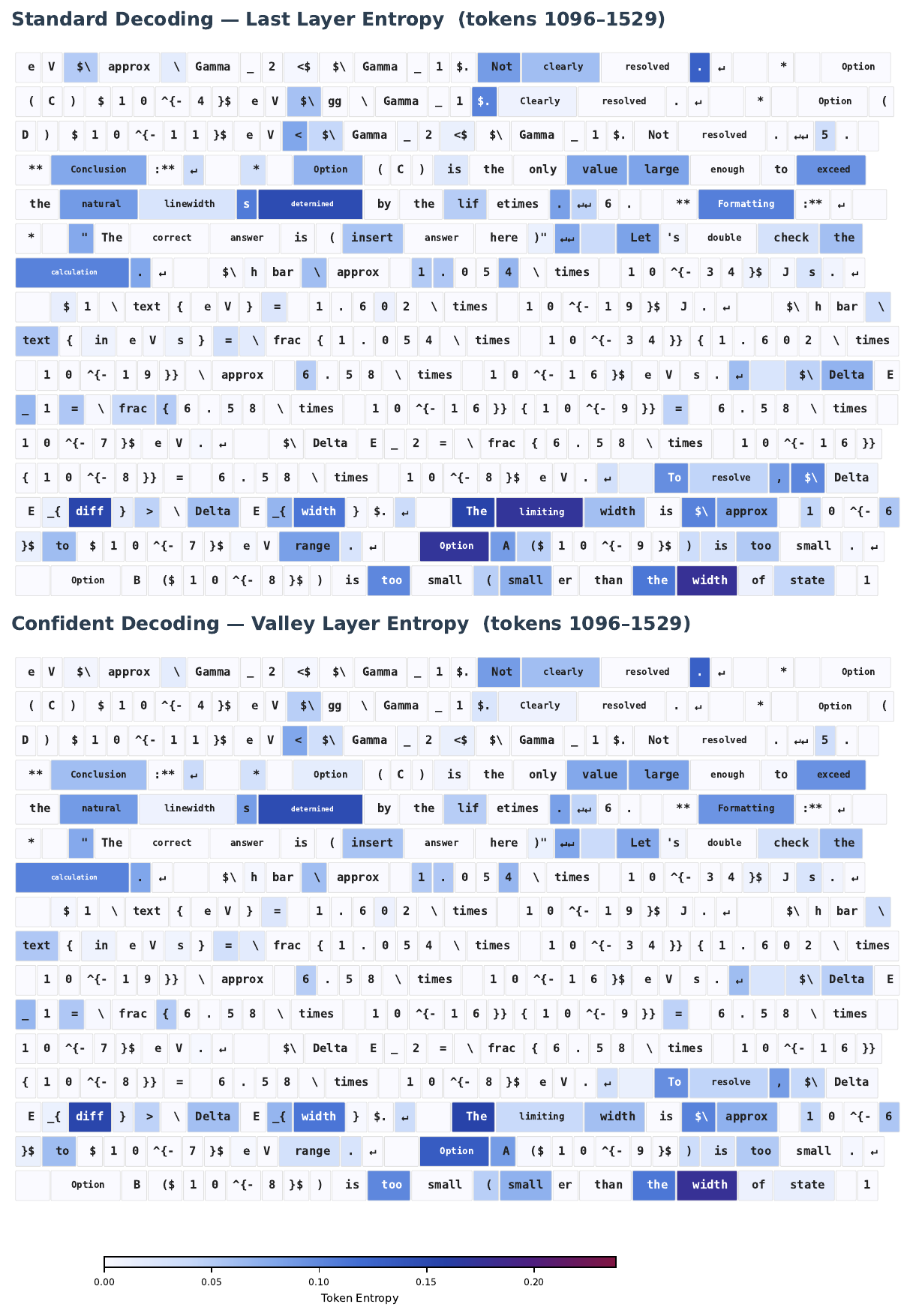}
    \vspace{-10pt}
    \caption{Visualization of token entropy (Part 4). For brevity, we omit the CoT following \cref{fig: visualize token entropy part 3} and preceding \cref{fig: visualize token entropy part 20}.}
    \label{fig: visualize token entropy part 20}
\end{figure}

\begin{figure}[!h]
    \centering
    \includegraphics[width=.95\linewidth]{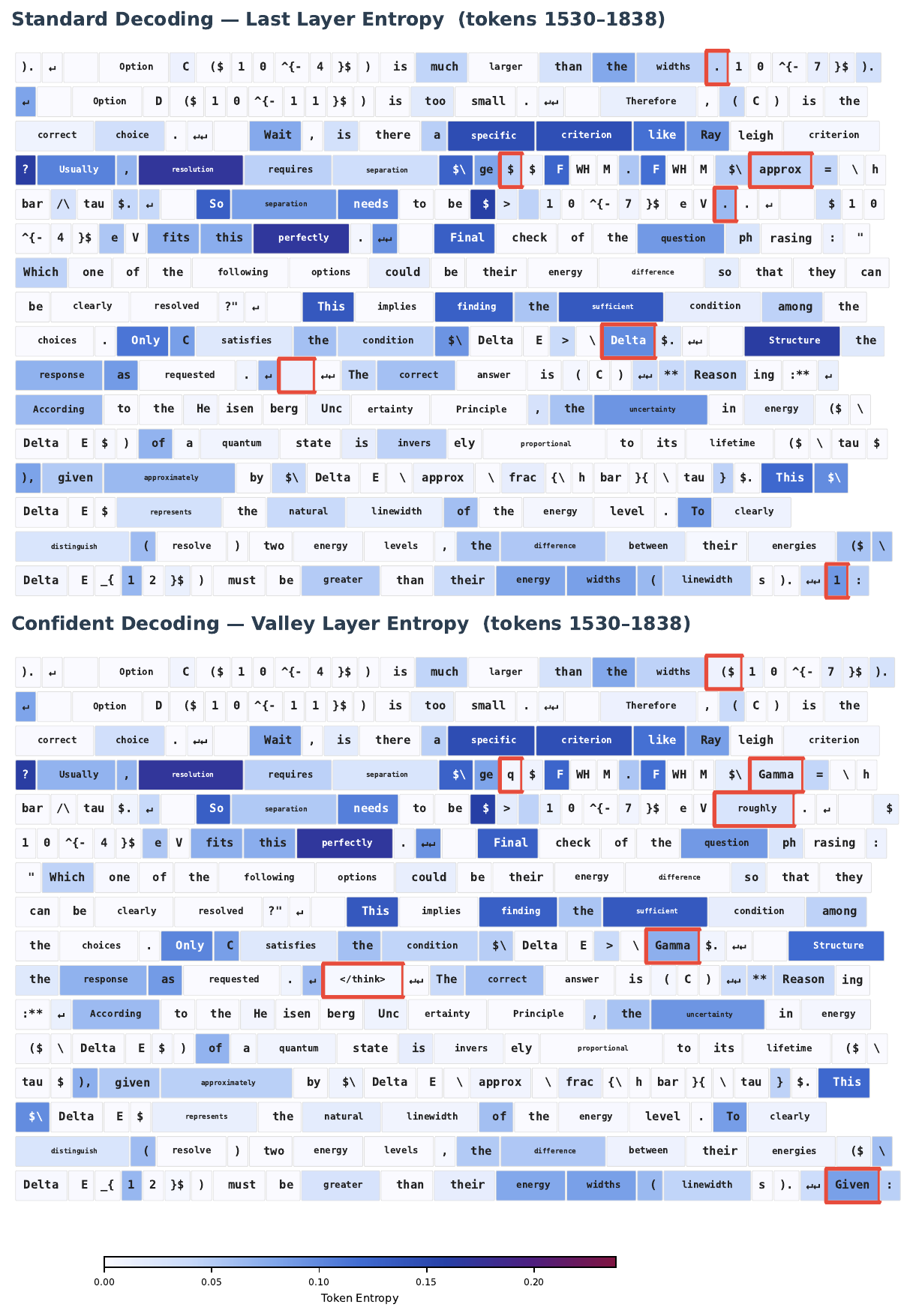}
    \vspace{-10pt}
    \caption{Visualization of token entropy (part 5).}
    \label{fig: visualize token entropy part 21}
\end{figure}

\begin{figure}[!h]
    \centering
    \includegraphics[width=.95\linewidth]{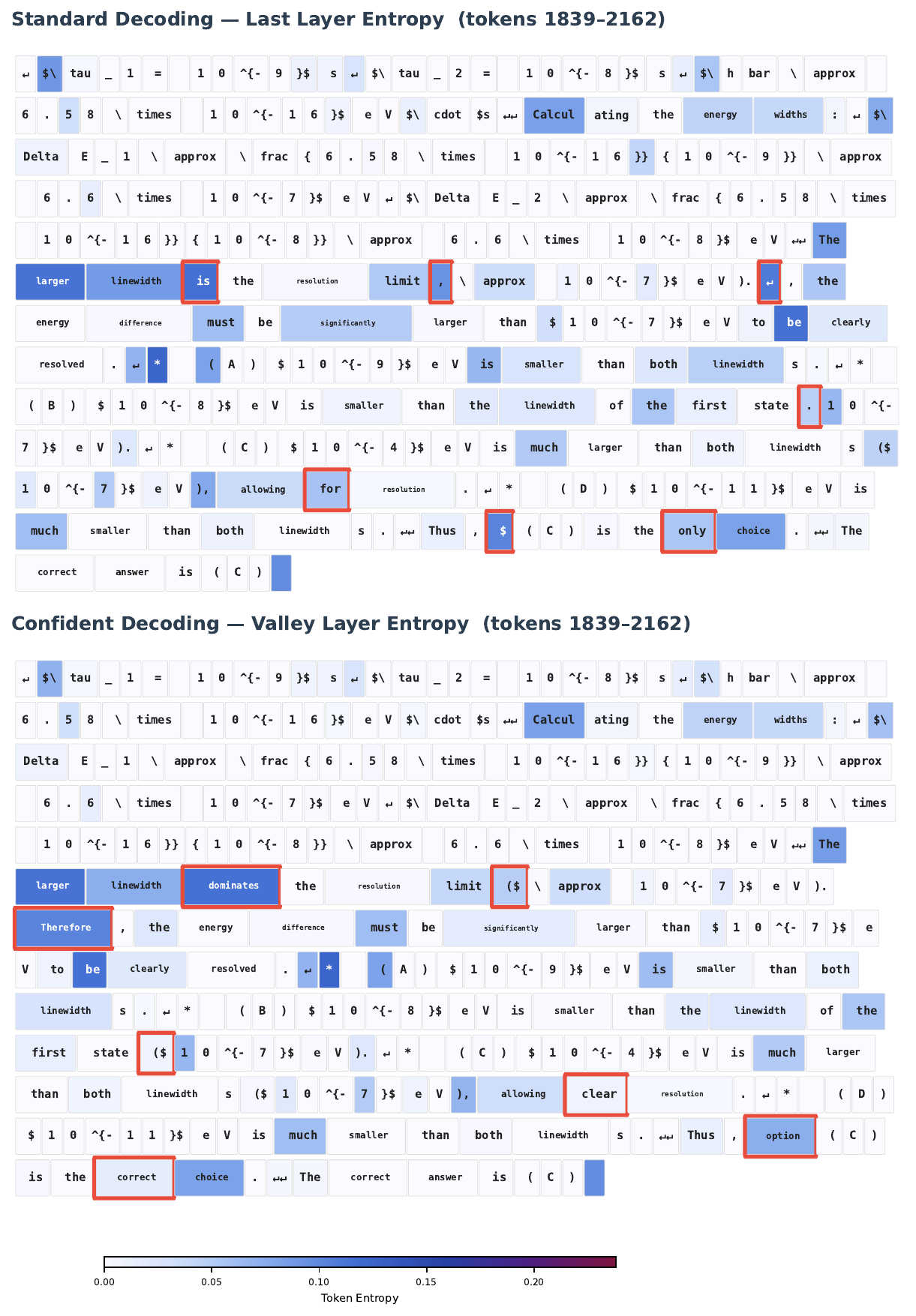}
    \vspace{-10pt}
    \caption{Visualization of token entropy (part 6).}
    \label{fig: visualize token entropy part 22}
\end{figure}

\textit{Algorithm-Specific Hyperparameters ($K$ and $p$).} The lookback window $K$ specifies how many transformer layers the backward scan in the main algorithm can traverse before terminating, while $p$ governs whether the decoded token is taken from the entropy valley ($p$) or from the standard final layer ($1-p$). Throughout all main experiments, we generally fix $K{=}10$ and $p{=}1.0$. The choice of $K{=}10$ is driven primarily by algorithmic efficiency: empirically, across all six benchmarks and all model scales we evaluate, the entropy valley located by the backward scan rarely lies more than $5$ layers below the final layer (as shown in our entropy-trajectory visualizations in Figures~\ref{fig: visualize token entropy part 1}--\ref{fig: visualize token entropy part 22}). A window of $K{=}10$ therefore covers the vast majority of valleys with substantial slack, while keeping the per-token logit-computation overhead bounded. Setting $p{=}1.0$ corresponds to a fully deterministic valley-selection rule: whenever a valley is detected within the lookback window, we always read out the token from that layer. As shown both in our motivating analysis (Figure~\ref{fig: superiority}) and in the ablation reported below (Table~\ref{tab:ablation_p_temp}), deterministic valley selection consistently outperforms stochastic mixtures, indicating every additional draw from the perturbed final layer reintroduces precisely the alignment-tax bias that \textsc{Confident Decoding} is designed to bypass.

\textit{Sampling Parameters ($T$ and top-$p$).} For all benchmarks we use greedy-style decoding with temperature $T{=}0.0$ and top-$p{=}1.0$, which mirrors the default configuration used by the corresponding official evaluation repositories (GPQA-Diamond, HLE, LiveCodeBench, LongBench v2, Omni-MATH, Air-Bench 2024 and WritingBench). This choice serves two purposes. First, it makes our baselines comparable to community-reported scores. Second, it isolates the contribution of \textsc{Confident Decoding}: any performance change we observe must come from the layer at which the next-token distribution is read, not from sampling-induced diversity. The ablation in Table~\ref{tab:ablation_p_temp} further supports this design choice.

\textit{Benchmark-Specific Settings.} Table~\ref{tab:bench_settings} summarizes the prompt template and decoding configuration we use for each benchmark, all of which are inherited from the corresponding official codebase with only the model-serving backend swapped to a vLLM-hosted OpenAI-compatible endpoint:

\begin{tcolorbox}[breakable, colback=gray!5!white, colframe=gray!75!black]
\footnotesize

\subsubsection*{GPQA-Diamond \citep{rein2024gpqa}}
\textbf{System Prompt}: 

\textit{(none — zero-shot setting)}

\textbf{User Prompt Template}:
\begin{verbatim}
What is the correct answer to this question: {question}

Choices:
(A) {choice1}
(B) {choice2}
(C) {choice3}
(D) {choice4}

Format your response as follows: 'The correct answer is (insert answer here)'
\end{verbatim}
\textbf{Decoding Settings}: $T{=}0.0$, top-$p{=}1.0$, max tokens $32{,}768$.

\textbf{Grading}: Exact-match extraction of the answer letter from the model output.

\vspace{0.5em}
\hrule
\vspace{0.5em}

\subsubsection*{HLE — Humanity's Last Exam \citep{center2026benchmark}}
\textbf{System Prompt}:
\begin{verbatim}
Your response should be in the following format:
Explanation: {your explanation for your answer choice}
Answer: {your chosen answer}
Confidence: {your confidence score between 0% and 100% for your answer}
\end{verbatim}
\textbf{User Prompt}: 

The raw question text from the HLE dataset (supports multi-modal with optional image URLs).

\textbf{Decoding Settings}: $T{=}0.0$, top-$p{=}1.0$, max tokens $32{,}768$.

\textbf{Grading}: LLM-as-a-judge with \texttt{gpt-5.4} evaluating Explanation/Answer/Confidence against ground truth.

\vspace{0.5em}
\hrule
\vspace{0.5em}

\subsubsection*{LiveCodeBench v6 — Code Generation \citep{jainlivecodebench}}
\textbf{System Prompt}:
\begin{verbatim}
You are an expert Python programmer. You will be given a question (problem specification) and will 
generate a correct Python program that matches the specification and passes all tests.
\end{verbatim}
\textbf{User Prompt Template}:
\begin{verbatim}
### Question:
{question_content}

### Format:
{starter_code_message}

### Answer: (use the provided format with backticks)
\end{verbatim}
\textbf{Decoding Settings}: $T{=}0.0$, top-$p{=}1.0$, max tokens $32{,}768$, $n{=}1$ generation per problem.

\textbf{Grading}: Official functional-correctness evaluator (release \texttt{v6}) running the generated code against test cases.

\vspace{0.5em}
\hrule
\vspace{0.5em}

\subsubsection*{LongBench v2 \citep{bai2025longbench}}
\textbf{System Prompt}: 

\textit{(none — in-context only)}

\textbf{User Prompt Template}:
\begin{verbatim}
Please read the following text and answer the questions below.

<text>
$DOC$
</text>

What is the correct answer to this question: $Q$
Choices:
(A) $C_A$
(B) $C_B$
(C) $C_C$
(D) $C_D$

Please think step by step.
\end{verbatim}
\textbf{Decoding Settings}: $T{=}0.0$, top-$p{=}1.0$, CoT phase max tokens $32{,}768$, answer phase max tokens $16{,}384$. Long inputs truncated by head–tail concatenation to fit the model's context window (official truncation strategy).
\textbf{Grading}: Exact-match of the extracted answer letter.

\vspace{0.5em}
\hrule
\vspace{0.5em}

\subsubsection*{Omni-MATH \citep{gaoomni}}
\textbf{System Prompt}:
\begin{verbatim}
You are a helpful and harmless assistant. You should think step-by-step and put your final answer 
within \boxed{}.
\end{verbatim}
\textbf{User Prompt}: 

The problem statement (field \texttt{question} or \texttt{problem} in the dataset JSONL).

\textbf{Decoding Settings}: $T{=}0.0$, top-$p{=}1.0$, max tokens $32{,}768$, $n{=}1$.

\textbf{Grading}: Official rule-based math evaluator — \texttt{parse\_ground\_truth} + \texttt{run\_execute} + \texttt{math\_equal}.

\vspace{0.5em}
\hrule
\vspace{0.5em}

\subsubsection*{Air-Bench 2024 \citep{zeng2025air}}
\textbf{System Prompt}:
\begin{verbatim}
You are a helpful assistant.
\end{verbatim}
\textbf{User Prompt}: 

Question text sampled from the official taxonomy (5 samples per leaf category, \texttt{sample\_size}{=}5).

\textbf{Decoding Settings}: $T{=}0.0$, top-$p{=}1.0$, max tokens $512$.

\textbf{Grading}: Official judge prompts evaluated by \texttt{gpt-5.4} via LLM-as-a-judge; the judge outputs a short reasoning and a numerical score.

\vspace{0.5em}
\hrule
\vspace{0.5em}

\subsubsection*{WritingBench \citep{wu2026writingbench}}
\textbf{System Prompt}: \textit{(none — direct user query)}

\textbf{User Prompt}: Raw writing query from the official benchmark (1,000 queries spanning 6 domains: Academic \& Engineering, Finance \& Business, Politics \& Law, Literature \& Arts, Education, Advertising \& Marketing).

\textbf{Decoding Settings}: $T{=}0.7$, top-$p{=}0.8$, top-$k{=}20$, max tokens $16{,}000$. These generation parameters follow the official WritingBench leaderboard configuration.

\textbf{Scoring}: Two-stage pipeline: (1) candidate model generates a response; (2) a judge model (default: \texttt{gpt-5.4}) scores the response on a 1–10 scale against five instance-specific criteria, each requiring a score and a textual justification. The system prompt for the judge is \textit{``You are an expert evaluator with extensive experience in evaluating response of given query.''}, and the scoring prompt follows the official rubric with scoring bands 1–2 (critical deficiencies), 3–4 (noticeable shortcomings), 5–6 (adequate baseline), 7–8 (strong with minor refinements), and 9–10 (exceptional). The final WritingBench score is the mean over all criteria across all queries.

\end{tcolorbox}

\begin{table}[!h]
\centering
\caption{Benchmark-specific decoding and evaluation settings used in our experiments. All configurations follow the corresponding official repositories. ``Judge'' indicates the auxiliary model used for LLM-as-a-judge scoring, where applicable.}
\label{tab:bench_settings}
\resizebox{0.95\textwidth}{!}{
\begin{tabular}{lcccl}
\toprule
\textbf{Benchmark} & \textbf{Temperature} & \textbf{Max Tokens} & \textbf{Judge} & \textbf{Prompt Source} \\
\midrule
GPQA-Diamond     & 0.0 & 32{,}768 & --        & official zero-shot \\
HLE              & 0.0 & 32{,}768 & gpt-5.4   & official system prompt \\
LiveCodeBench v6 & 0.0 & 32{,}768 & --        & official code-generation template \\
LongBench v2     & 0.0 & 32{,}768 (CoT) & -- & official \texttt{0shot\_cot} \\
Omni-MATH        & 0.0 & 32{,}768 & rule-based & official math system prompt \\
Air-Bench 2024   & 0.0 & 512      & gpt-5.4   & official taxonomy + judge prompts \\
WritingBench     & 0.7 & 16{,}000 & gpt-5.4   & official taxonomy + judge prompts \\
\bottomrule
\end{tabular}
}
\end{table}

\textit{Sensitivity Analysis on $p$ and $T$.} We ablate the two most influential decoding knobs—the valley-selection probability $p$ and the temperature $T$—on GPQA-Diamond using \textbf{Qwen3.5-35B-A3B}. Results are reported in Table~\ref{tab:ablation_p_temp}.

\begin{table}[!h]
\centering
\caption{Ablation of $p$ and temperature on GPQA-Diamond accuracy with Qwen3.5-35B-A3B. The default setting ($T{=}0.0$, $p{=}1.0$) achieves the best accuracy. Top block: vary $T$ at $p{=}1.0$; bottom block: vary $p$ at $T{=}0.0$ (where $p{=}0.0$ recovers standard greedy decoding).}
\label{tab:ablation_p_temp}
\begin{tabular}{ccc}
\toprule
\textbf{Temperature $T$} & \textbf{$p$} & \textbf{GPQA-Diamond Acc.\ (\%)} \\
\midrule
0.0 & 1.0 & \textbf{82.8} \\
0.1 & 1.0 & 80.0 \\
0.4 & 1.0 & 81.8 \\
0.7 & 1.0 & 81.3 \\
1.0 & 1.0 & 80.8 \\
\midrule
0.0 & 0.8 & 81.8 \\
0.0 & 0.6 & 80.0 \\
0.0 & 0.4 & 79.8 \\
0.0 & 0.2 & 78.3 \\
0.0 & 0.0 & 76.3 \\
\bottomrule
\end{tabular}
\end{table}

Two trends emerge clearly. First, fixing $p{=}1.0$ and increasing the temperature from $0.0$ to $1.0$ degrades accuracy almost monotonically (from $82.8\%$ to $80.8\%$). This indicates that, once the decoding layer is correctly chosen, additional stochasticity in the softmax draws back in the very alignment-tax noise that \textsc{Confident Decoding} suppresses, and yields no benefit. Second, fixing $T{=}0.0$ and decreasing $p$ from $1.0$ down to $0.0$ produces a strict, monotone collapse from $82.8\%$ to $76.3\%$. Note that the lower endpoint $p{=}0.0$ exactly recovers standard greedy decoding from the final layer, and our deterministic valley-selection setting ($p{=}1.0$) improves over it by $+6.5$ absolute points. Intermediate values of $p$ smoothly interpolate between the two regimes, which directly mirrors the theoretical picture in Section~\ref{subsec: alignment-tax}: each unit of probability mass placed back on the perturbed final layer reintroduces a proportional fraction of the KL penalty $\mathcal{D}_{\text{KL}}(P_{\text{logic}} \| P_{\text{align}})$. Consequently, the deterministic-valley regime ($p{=}1.0$, $T{=}0.0$) is not merely a tuned best operating point—it is the configuration the theory predicts, and we adopt it as our default in all main experiments.

\section{Three-Phase Structure on More Models}
\label{app:9b-replication}

The main text presents layer-wise dynamics (Figure~\ref{fig: layer wise}) and per-token entropy partitioning (Figure~\ref{fig: theory prove}) on the Qwen3.5-35B-A3B backbone. To verify that the three-phase structure and the per-token Phase~III heterogeneity are not artifacts of the MoE architecture or model scale, we replicate both analyses on Qwen3.5-9B---a compact hybrid model with $L{=}32$ layers (24 DeltaNet + 8 full-attention) and no sparse expert routing.

\subsection{Layer-wise Dynamics}
\label{app:9b-layerwise}

\begin{figure}[!t]
    \centering
    \begin{minipage}[t]{0.47\linewidth}
        \centering
        \includegraphics[width=\linewidth]{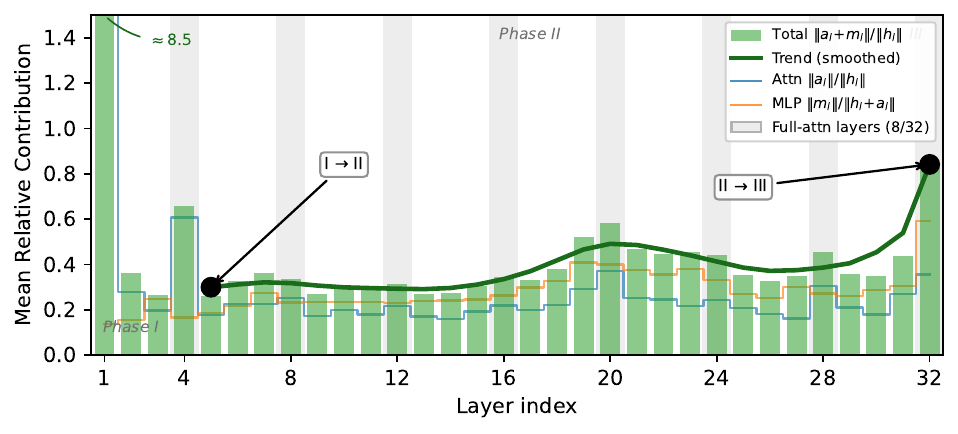}
        \subcaption{Relative Contribution Norm}
        \label{fig: layer wise norm 9b}
    \end{minipage}
    \hfill
    \begin{minipage}[t]{0.47\linewidth}
        \centering
        \includegraphics[width=\linewidth]{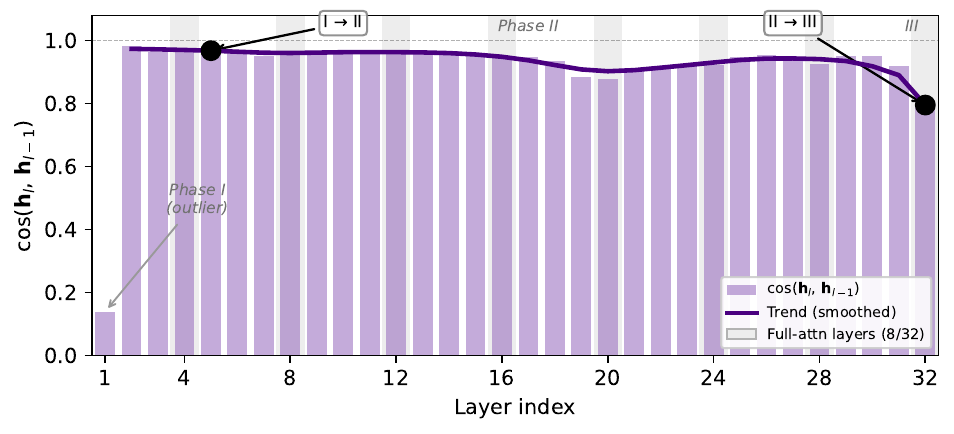}
        \subcaption{Residual I/O Cosine Similarity}
        \label{fig: layer wise cossim 9b}
    \end{minipage}
    \caption{Layer-wise dynamics of Qwen3.5-9B-Base on GSM8K (cf.\ Figure~\ref{fig: layer wise} for 35B-A3B). Gray bands mark the 8 full-attention layers ($l \in \{4,8,12,16,20,24,28,32\}$).
    \textbf{(a)} Norm Ratio: Phase~I overwrite at $l{=}1$ (${\approx}8.5$), stable Phase~II ($0.26$--$0.58$), Phase~III resurgence at $l{=}32$ (${\approx}0.84$).
    \textbf{(b)} IO-CosSim: high fidelity throughout Phase~II ($0.88$--$0.97$), dropping to $0.80$ at $l{=}32$. The three-phase structure is qualitatively identical to the 35B-A3B result.}
    \label{fig: layer wise 9b}
\end{figure}

Figure~\ref{fig: layer wise 9b} reports the Relative Contribution Norm and Residual I/O Cosine Similarity for Qwen3.5-9B-Base on GSM8K, mirroring the 35B-A3B analysis in Figure~\ref{fig: layer wise}. The three-phase progression is clearly preserved:
\begin{itemize}[leftmargin=15pt, topsep=2pt, itemsep=2pt]
    \item \textbf{Phase~I} ($l{=}1$): Norm Ratio${\approx}8.5$ with IO-CosSim${\approx}0.14$, confirming that the first layer effectively overwrites the token embedding.
    \item \textbf{Phase~II} ($5 \le l \le 31$): Norm Ratio stabilises at $0.26$--$0.58$ and IO-CosSim remains high ($0.88$--$0.97$), consistent with incremental, directionally faithful refinement.
    \item \textbf{Phase~III} ($l{=}32$): Norm Ratio resurges to ${\approx}0.84$ (roughly $2{\times}$ the Phase~II plateau) and IO-CosSim drops to ${\approx}0.80$---the largest directional deflection in the late regime, matching the pattern observed in 35B-A3B.
\end{itemize}

\subsection{Per-Token Entropy Partitioning}
\label{app:9b-entropy}

\begin{figure}[!t]
    \centering
    \begin{minipage}[t]{0.48\linewidth}
        \centering
        \includegraphics[width=\linewidth]{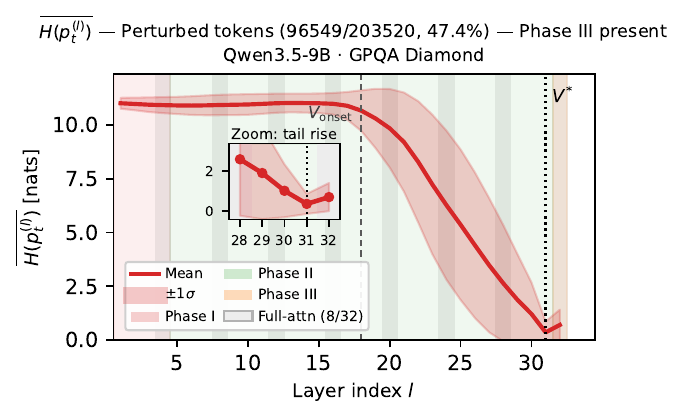}
        \subcaption{Perturbed tokens --- Phase~III present}
        \label{fig: theory rising 9b}
    \end{minipage}
    \hfill
    \begin{minipage}[t]{0.48\linewidth}
        \centering
        \includegraphics[width=\linewidth]{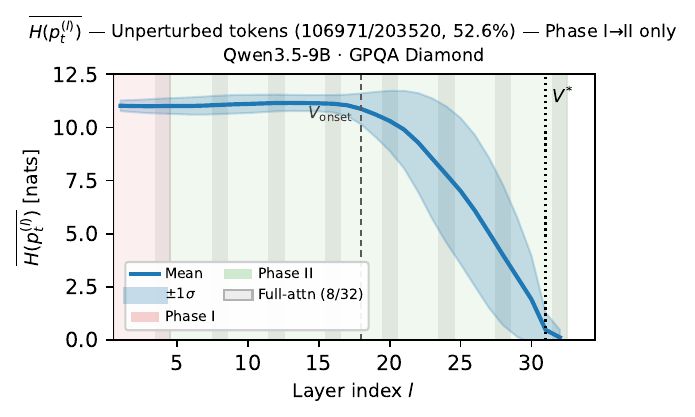}
        \subcaption{Unperturbed tokens --- no Phase~III}
        \label{fig: theory falling 9b}
    \end{minipage}
    \caption{Mean logit-lens entropy $\overline{H(p^{(l)}_t)}$ per layer for Qwen3.5-9B on GPQA Diamond (cf.\ Figure~\ref{fig: theory prove} for 35B-A3B). Tokens are partitioned by whether the final layer ($l{=}32$) raises entropy relative to $l{=}31$.
    \textbf{(a)} Perturbed tokens (47.4\%, mean $\Delta H{=}{+}0.34$ nats) show a Phase~III entropy rise at the final full-attention layer, mirroring the 35B-A3B pattern but at a substantially higher rate (47.4\% vs.\ 16.2\%), consistent with the compressed corridor ($L{=}32$ vs.\ $L{=}40$).
    \textbf{(b)} Unperturbed tokens (52.6\%) continue their Phase~II refinement through $l{=}32$ without disruption. The backward scan naturally selects $l{=}32$ for these tokens, recovering standard decoding.}
    \label{fig: theory prove 9b}
\end{figure}

Figure~\ref{fig: theory prove 9b} replicates the per-token entropy partitioning from Figure~\ref{fig: theory prove} on Qwen3.5-9B using GPQA Diamond ($N{=}50$ prompts, 4{,}096 generated tokens per prompt, 203{,}520 tokens total). Following the same protocol, each generated token is classified as \emph{perturbed} (entropy rises at the final layer: $H^{(32)}_t > H^{(31)}_t$) or \emph{unperturbed} (entropy does not rise). On the 9B backbone, 47.4\% of tokens are perturbed (mean $\Delta H {=} {+}0.34$ nats), compared with only 16.2\% on 35B-A3B. This substantially higher perturbation rate is consistent with the compressed refinement corridor ($L{=}32$ vs.\ $L{=}40$) discussed in Appendix~\ref{app:degradation}: fewer layers complete the Phase~II refinement, leaving the final full-attention layer more likely to overshoot. Despite this quantitative difference, the qualitative three-phase pattern is preserved, confirming that per-token Phase~III heterogeneity is a general phenomenon across model scales.

\section{Token-Level Substitution Analysis}
\label{app:token-substitution}

To shed light on \emph{what} Confident Decoding actually changes at the token level, we inspect the tokens that are substituted when the entropy-trough layer disagrees with the final layer. Table~\ref{tab:token-substitution-ranking} lists the 20 most frequently substituted tokens for both the Base and Instruct variants of Qwen3.5-35B-A3B on GPQA-Diamond, together with a category breakdown of all substitution events.

\begin{table*}[t]
\centering
\caption{Top-20 most frequently substituted tokens when Confident Decoding disagrees with Standard Decoding, comparing Base and Instruct variants of Qwen3.5-35B-A3B on GPQA-Diamond. \textit{Standard} columns show the tokens that the final layer (layer~40) would emit; \textit{Confident} columns show what the entropy-trough layer selects instead. Percentages denote the fraction of all substitution events in each model. Category breakdowns (Content / Function / Punctuation) are summarized at the bottom. Base: 956 substitutions out of 40,526 tokens (2.4\%); Instruct: 939 out of 36,111 (2.6\%).}
\label{tab:token-substitution-ranking}
\small
\setlength{\tabcolsep}{3.5pt}
\begin{tabular}{@{}r l r  r l r  r l r  r l r@{}}
\toprule
& \multicolumn{5}{c}{\textbf{Base Model}} & & \multicolumn{5}{c}{\textbf{Instruct Model}} \\
\cmidrule(lr){2-6} \cmidrule(l){8-12}
& \multicolumn{2}{c}{Standard (Last Layer)} & & \multicolumn{2}{c}{Confident (Trough)} &
& \multicolumn{2}{c}{Standard (Last Layer)} & & \multicolumn{2}{c}{Confident (Trough)} \\
\cmidrule(lr){2-3} \cmidrule(lr){5-6} \cmidrule(lr){8-9} \cmidrule(l){11-12}
1 & \texttt{.} & 4.8\% & & \texttt{Actually} & 4.1\% & & \texttt{.} & 7.7\% & & \texttt{(\$} & 6.7\% \\
2 & \texttt{the} & 4.4\% & & \texttt{\textless{}\textbar{}box\_end\textbar{}\textgreater{}} & 2.5\% & & \texttt{\textbackslash\{\}n} & 4.2\% & & \texttt{carb} & 2.1\% \\
3 & \texttt{,} & 3.1\% & & \texttt{careful} & 2.3\% & & \texttt{the} & 4.2\% & & \texttt{ring} & 1.8\% \\
4 & \texttt{is} & 2.9\% & & \texttt{perpendicular} & 1.9\% & & \texttt{\$} & 3.5\% & & \texttt{\$,} & 1.7\% \\
5 & \texttt{(space)} & 2.7\% & & \texttt{carb} & 1.8\% & & \texttt{is} & 3.5\% & & \texttt{\$\textbackslash\{\}} & 1.6\% \\
6 & \texttt{k} & 2.5\% & & \texttt{phenotype} & 1.7\% & & \texttt{\$} & 3.0\% & & \texttt{Structure} & 1.6\% \\
7 & \texttt{g} & 2.0\% & & \texttt{fusion} & 1.7\% & & \texttt{(space)} & 2.1\% & & \texttt{\$.} & 1.4\% \\
8 & \texttt{(} & 1.9\% & & \texttt{maybe} & 1.5\% & & \texttt{(} & 2.1\% & & \texttt{\$(} & 1.3\% \\
9 & \texttt{:} & 1.9\% & & \texttt{fused} & 1.5\% & & \texttt{,} & 2.0\% & & \texttt{option} & 1.1\% \\
10 & \texttt{a} & 1.8\% & & \texttt{indeed} & 1.4\% & & \texttt{\$\textbackslash\{\}} & 1.7\% & & \texttt{Sometimes} & 1.0\% \\
11 & \texttt{So} & 1.8\% & & \texttt{actually} & 1.3\% & & \texttt{).} & 1.5\% & & \texttt{methyl} & 0.9\% \\
12 & \texttt{to} & 1.7\% & & \texttt{each} & 1.3\% & & \texttt{?} & 1.5\% & & \texttt{**+} & 0.9\% \\
13 & \texttt{with} & 1.4\% & & \texttt{neither} & 1.2\% & & \texttt{\textbackslash\{\}} & 1.4\% & & \texttt{\textbackslash\{\}} & 0.9\% \\
14 & \texttt{The} & 1.3\% & & \texttt{an} & 1.0\% & & \texttt{The} & 1.3\% & & \texttt{normalization} & 0.7\% \\
15 & \texttt{that} & 1.3\% & & \texttt{mutating} & 1.0\% & & \texttt{a} & 1.3\% & & \texttt{definitely} & 0.7\% \\
16 & \texttt{it} & 1.2\% & & \texttt{exactly} & 0.8\% & & \texttt{=} & 1.2\% & & \texttt{Corey} & 0.6\% \\
17 & \texttt{G} & 1.2\% & & \texttt{oxygen} & 0.8\% & & \texttt{So} & 1.2\% & & \texttt{an} & 0.6\% \\
18 & \texttt{)} & 1.0\% & & \texttt{sqrt} & 0.8\% & & \texttt{:} & 1.1\% & & \texttt{?).} & 0.6\% \\
19 & \texttt{But} & 1.0\% & & \texttt{acting} & 0.8\% & & \texttt{to} & 1.0\% & & \texttt{+} & 0.6\% \\
20 & \texttt{are} & 1.0\% & & \texttt{sometimes} & 0.7\% & & \texttt{Carb} & 1.0\% & & \texttt{'s} & 0.6\% \\
\midrule
\multicolumn{12}{@{}l}{\textit{Category breakdown of all substituted tokens (\%):}} \\
& \quad Content & 29\% & & & 77\% & & & 26\% & & & 60\% \\
& \quad Function & 39\% & & & 10\% & & & 27\% & & & 6\% \\
& \quad Punctuation & 22\% & & & 6\% & & & 43\% & & & 26\% \\
& \quad Other & 10\% & & & 7\% & & & 4\% & & & 8\% \\
\bottomrule
\end{tabular}
\end{table*}

\section{More Details of the Baselines}
\label{app:baselines}

We compare Confident Decoding against two representative contrastive decoding baselines: Decoding by Contrasting Layers (DoLa)~\citep{chuangdola} and Self Logits Evolution Decoding (SLED)~\citep{zhang2024sled}. Both methods were originally developed and evaluated on standard dense Transformers (e.g., LLaMA-family models), and their official implementations rely on layer-indexing conventions and residual-stream access patterns specific to homogeneous dense architectures. Consequently, neither can be directly applied to modern hybrid or MoE backbones such as Qwen3.5-35B-A3B, where interleaved DeltaNet/full-attention layers and sparse expert routing fundamentally alter the layer-wise representation geometry.

To enable a fair comparison, we re-implemented both DoLa and SLED within the same inference framework used for Confident Decoding, adapting them to operate correctly on the hybrid MoE architecture. We first validated our re-implementations on LLaMA-family models, confirming that they reproduce the originally reported results. Table~\ref{tab:baseline_comparison} reports the results on Qwen3.5-35B-A3B. Both contrastive baselines yield modest improvements over last layer decoding, but their gains fall substantially short of those achieved by Confident Decoding.

\begin{table}[htbp]
\centering
\caption{Comparison of contrastive decoding baselines on Qwen3.5-35B-A3B. DoLa and SLED are re-implemented within our inference framework to support the hybrid MoE architecture. Both contrastive baselines yield smaller gains than Confident Decoding on this backbone.}
\label{tab:baseline_comparison}
\resizebox{0.9\textwidth}{!}{
\begin{tabular}{lccccccc}
\toprule
\textbf{Method} & \textbf{GPQA-D} & \textbf{HLE} & \textbf{LCB-v6} & \textbf{LongBench-v2} & \textbf{Omni-Math} & \textbf{Air-Bench} & \textbf{WritingBench} \\ \midrule
Last Layer            & 76.3 & 7.1  & 70.0 & \textbf{63.9} & 72.6 & 51.0 & 65.3 \\
DoLa              & 77.3 & 7.8  & 70.9 & 63.4 & 72.4 & 52.0 & 65.2 \\
SLED              & 78.8 & 7.4  & 71.7 & 63.2 & 72.1 & 53.0 & 65.3 \\
Confident (Ours)  & \textbf{82.8} & \textbf{9.5} & \textbf{75.1} & 63.7 & \textbf{72.7} & \textbf{56.0} & \textbf{65.4} \\
\bottomrule
\end{tabular}
}
\end{table}

The limited effectiveness of contrastive baselines on hybrid MoE architectures is structurally predictable. For example, DoLa dynamically selects a premature layer that maximally diverges from the final layer (measured by Jensen--Shannon divergence) and decodes from the contrastive log-probability difference between the two, implicitly assuming that early and late layers occupy a representationally comparable subspace differing only in depth of reasoning. On hybrid backbones, the interleaving of DeltaNet and full-attention layers introduces discontinuous shifts in representation geometry (as formalized by the $\epsilon^{(l)}_{\text{type}}$ term in our degradation analysis), breaking this subspace-comparability assumption. The contrastive signal is therefore diluted by architectural noise, reducing its ability to isolate the intended factuality gradient. SLED applies gradient-style corrections based on the evolution of logits across layers, requiring the logit trajectory to be locally smooth and monotonically informative. The non-monotonic entropy profile induced by layer-type alternation and expert switching corrupts the logit-evolution signal, limiting the effectiveness of the corrections.

Both baselines share a common structural limitation: they assume representational homogeneity across the layers being contrasted---a condition that modern hybrid and MoE architectures do not satisfy. Confident Decoding mitigates this limitation through a fundamentally different evaluation strategy: rather than contrasting logit distributions across layers (which requires representational comparability between the contrasted pair), it independently evaluates the predictive entropy at each candidate layer and selects the one with minimal uncertainty. Although it relies on the same unembedding projection $W_U$ and is therefore not immune to probe mismatch (as analyzed in the degradation analysis below), the per-layer independent evaluation avoids amplifying structured noise through cross-layer subtraction, making it less sensitive to the heterogeneous geometry of modern architectures.

\section{Degradation Analysis}
\label{app:degradation}

Table~\ref{tab:main_performance_all} reveals that Confident Decoding does not uniformly benefit all backbones. MoE models (e.g., Qwen3.5-35B-A3B) and larger Transformer-based models (e.g., Qwen3.5-27B, gpt-oss-20b) generally benefit or maintain performance across benchmarks, while the compact hybrid Qwen3.5-9B presents a mixed picture: it suffers notable regressions on certain benchmarks (e.g., GPQA-D: $64.6 \to 62.1$, Omni-MATH: $49.1 \to 47.1$) yet simultaneously achieves clear improvements on others (LCB-v6: $41.1 \to 47.7$, Air-Bench: $53.0 \to 56.0$, HLE: $5.2 \to 6.1$). This section analyzes the degradation through two interacting factors---\emph{architectural heterogeneity} and \emph{model depth}---and discusses what structural properties make certain backbones more amenable to Confident Decoding.

\textbf{Factor~I: Hybrid architecture compresses the refinement corridor.}
Qwen3.5 family employs a hybrid design. For example, Qwen3.5-9B interleaves $24$ linear-attention (DeltaNet) layers with $8$ full-attention layers ($L{=}32$). DeltaNet layers perform recurrent-style state updates that produce representations with a stronger ``process state'' character, while full-attention layers periodically reorganize global context. Because adjacent layers may belong to fundamentally different computational paradigms, the late \emph{refinement corridor}---the band of layers where the model's token-level prediction monotonically sharpens---is substantially compressed. Consequently, even a single-layer rollback in the backward scan can cross the decision boundary between the final full-attention consolidation layer and a pre-final DeltaNet state that has not yet completed global-context integration.

The root mechanism is a \emph{probe mismatch} induced by layer-type alternation. The backward scan evaluates entropy by projecting each intermediate hidden state through the shared final unembedding matrix $W_U$---a projection that is well-calibrated only when the probed layer's representation lies in roughly the same geometric subspace as the final layer's. In the hybrid architecture, however, successive layers alternate between linear-attention recurrence and full-attention recomposition, causing the representation geometry to shift at each type boundary. The effective probe error $\epsilon^{(l)}$ therefore acquires a structured, layer-type-dependent component. Decomposing the projection noise $\epsilon^{(l)}$ from Section~\ref{subsec: backward-optimality} into a smooth baseline term and a layer-type term:
\begin{equation}
\hat{H}(l) = H^*(l) + \epsilon^{(l)}_{\text{probe}} + \epsilon^{(l)}_{\text{type}} + \eta^{(l)},
\end{equation}
where $\epsilon^{(l)}_{\text{type}}$ exhibits discontinuous jumps at DeltaNet$\leftrightarrow$full-attention transitions. When this structured noise dominates the true entropy gradient $\Delta H^*(l)$, the observed entropy valley $\arg\min_l \hat{H}(l)$ can deviate from the true semantic optimum $\arg\min_l H^*(l)$, causing the backward scan to commit tokens to pre-convergent representations. Notably, this probe mismatch is \emph{task-dependent}: on benchmarks where the entropy valley signal is strong and concentrated in the final few layers (e.g., code generation in LCB-v6), the structured noise is insufficient to derail the scan, and even the 9B backbone achieves a clear gain ($+6.6$). The degradation concentrates on tasks such as GPQA-D and Omni-MATH, where the entropy gradient in the refinement corridor is shallower and more susceptible to being masked by $\epsilon^{(l)}_{\text{type}}$.

\textbf{Factor~II: Model depth widens the refinement corridor.}
Network depth modulates the degradation through a straightforward mechanism: deeper networks allocate more consecutive layers of the same type to the late refinement zone, widening the corridor over which $|\Delta H^*(l)| \gg |\Delta \epsilon^{(l)}|$ holds and providing a larger margin for the backward scan to locate a reliable valley. Qwen3.5-9B ($L{=}32$) has only $8$ full-attention layers spread across the entire stack, leaving the final refinement zone with very few homogeneous layers. By contrast, Qwen3.5-35B-A3B ($L{=}40$) and Qwen3.5-27B have substantially deeper stacks, and the wider corridor trivially satisfies the local-monotonicity precondition. This is consistent with the data: Qwen3.5-27B exhibits no degradation on GPQA-D ($80.0 \to 80.0$) and gains $+10.1$ on LCB-v6. We note that depth and architecture are partially confounded in our model suite---the deeper models also differ in parameter count and expert routing---so we cannot fully disentangle the contribution of depth alone from other architectural factors. Nonetheless, the pattern is consistent: shallower networks with narrower corridors are more vulnerable to probe-error-induced misselection.

\textbf{Why MoE architectures appear robust.}
Qwen3.5-A series---an MoE backbone that also employs hybrid attention---achieves significant Confident Decoding gains. Two properties may contribute beyond their depth. First, sparse expert routing concentrates task-relevant updates in specialized sub-networks, which may amplify the true refinement signal $|\Delta H^*(l)|$ in the late corridor and make it easier to dominate probe noise. Second, because each token activates only a small subset of experts, the effective representation trajectory across adjacent layers is smoother than in a dense hybrid backbone of equivalent depth, potentially reducing the type-switching probe error $\epsilon^{(l)}_{\text{type}}$. However, we emphasize that in our current experimental setup, depth and MoE routing are confounded: disentangling their individual contributions would require comparing models that differ only in one of these axes, which we leave to future work.

\textbf{Scope delimitation.}
The degradation on Qwen3.5-9B does not invalidate the entropy-valley hypothesis. Rather, the failure arises when deterministic valley selection ($p{=}1.0$) is applied to a backbone whose refinement corridor is too narrow to tolerate universal rollback on certain task types. The core precondition for Confident Decoding---as formalized in the dominance condition of Theorem~1 ($\Delta \eta^{(l)} > 2\epsilon_{\max}$)---is that the late refinement corridor must be sufficiently wide for the true entropy gradient to dominate the total probe error: $|\Delta H^*(l)| \gg |\Delta \epsilon^{(l)}|$, where $\epsilon^{(l)} = \epsilon^{(l)}_{\text{probe}} + \epsilon^{(l)}_{\text{type}}$. This condition is more readily satisfied on deeper networks, but can be violated on compact hybrid architectures where layer-type alternation introduces structured probe noise and limited depth compresses the safe corridor.

We also note that even models for which Confident Decoding is broadly beneficial can exhibit task-specific regressions. For instance, Qwen3.5-27B---a deeper model that gains strongly on reasoning benchmarks---shows a $-3.0$ drop on Air-Bench ($67.0 \to 64.0$). This suggests that the alignment-tax mechanism and the probe-mismatch mechanism can interact differently depending on the task's reliance on final-layer formatting versus intermediate-layer reasoning. A complete characterization of these per-task dynamics is an important direction for future work.

These findings delineate two axes---\emph{architectural homogeneity} and \emph{model depth}---along which the applicability of Confident Decoding varies. 

\begin{table}[t]
\centering
\caption{Full results for last layer decoding and confident decoding across various model families.}
\label{tab:main_performance_all}
\resizebox{0.95\textwidth}{!}{ 
\begin{tabular}{llccccccc}
\toprule
\textbf{Model} & \textbf{Method} & \textbf{GPQA-D} & \textbf{HLE} & \textbf{LCB-v6} & \textbf{LongBench-v2} & \textbf{Omni-Math} & \textbf{Air-Bench} & \textbf{WritingBench} \\ \midrule
\multirow{2}{*}{Qwen3.5-0.8B}         & Last Layer   & 0.0  & 0.0 & 0.0  & 0.0  & 0.0  & 52.0 & 10.0 \\
                                      & Confident & 0.0  & 0.0 & 0.0  & 0.0  & 0.0  & 54.0 & 10.1 \\ \addlinespace
\multirow{2}{*}{Qwen3.5-9B}           & Last Layer    & 64.6 & 5.2 & 41.1 & 62.0 & 49.1 & 53.0 & 60.7 \\
                                      & Confident & 62.1 & 6.1 & 47.7 & 60.5 & 47.1 & 56.0 & 59.5 \\ \addlinespace
\multirow{2}{*}{Qwen3.5-9B-Base}      & Last Layer    & 67.7 & 5.4 & 49.1 & 54.4 & 54.7 & 76.0 & 61.1 \\
                                      & Confident & 65.7 & 6.3 & 50.5 & 56.1 & 53.0 & 76.0 & 63.0 \\ \addlinespace
\multirow{2}{*}{Qwen3.5-27B}          & Last Layer    & 80.0 & 16.0 & 63.7 & 63.7 & 76.1 & 67.0 & 66.7 \\
                                      & Confident & 80.0 & 17.0 & 73.8 & 64.0 & 76.1 & 64.0 & 66.9 \\ \addlinespace
\multirow{2}{*}{Qwen3.5-35B-A3B}      & Last Layer    & 76.3 & 7.1 & 70.0 & 63.9 & 72.6 & 51.0 & 65.3 \\
                                      & Confident & 82.8 & 9.5 & 75.1 & 63.7 & 72.7 & 56.0 & 65.4 \\ \addlinespace
\multirow{2}{*}{Qwen3.5-122B-A10B}    & Last Layer    & 83.3 & 14.7 & 76.8 & 66.4 & 78.3 & 66.0 & 72.6 \\
                                      & Confident & 85.4 & 16.5 & 79.2 & 66.7 & 78.5 & 67.0 & 72.7 \\ \addlinespace
\multirow{2}{*}{Gemma-4-31B}          & Last Layer    & 76.8 & 9.2 & 75.1 & 58.6 & 68.3 & 74.0 & 64.2 \\
                                      & Confident & 80.8 & 11.0 & 78.3 & 59.1 & 69.0 & 76.0 & 64.4 \\ \addlinespace
\multirow{2}{*}{gpt-oss-20b}          & Last Layer    & 57.1 & 5.5 & 77.4 & 44.6 & 60.9 & 86.0 & 54.4 \\
                                      & Confident & 60.1 & 6.4 & 78.0 & 46.3 & 61.6 & 89.0 & 54.9 \\ \addlinespace
\multirow{2}{*}{gpt-oss-120B}         & Last Layer    & 76.3 & 10.8 & 83.1 & 54.7 & 70.2 & 88.0 & 67.5 \\
                                      & Confident & 80.8 & 12.6 & 84.5 & 55.6 & 71.0 & 90.0 & 67.8 \\ \bottomrule
\end{tabular}
}
\end{table}

\end{document}